\documentclass{article}

\PassOptionsToPackage{numbers, compress}{natbib}

\usepackage[preprint]{neurips_2024}

\usepackage[utf8]{inputenc} 
\usepackage[T1]{fontenc}    
\usepackage{hyperref}       
\usepackage{url}            
\usepackage{booktabs}       
\usepackage{amsfonts}       
\usepackage{nicefrac}       
\usepackage{microtype}      
\usepackage{xcolor}         

\usepackage{amsmath}
\usepackage{algorithmic}
\usepackage{algorithm}
\usepackage{graphicx}

\usepackage{bbding}
\usepackage{pifont}
\usepackage{wasysym}

\usepackage{subfigure}
\usepackage{wrapfig}
\usepackage{enumitem}
\usepackage{booktabs}

\usepackage{mathtools}
\usepackage{amssymb}
\usepackage{amsthm}

\usepackage{footnote}
\usepackage{makecell}
\makesavenoteenv{tabular}
\usepackage{multirow}
\usepackage{xcolor}

\usepackage[font=small]{caption}

\newcommand{\T}{\mathcal{T}}
\newcommand{\K}{\mathcal{K}}
\newcommand{\R}{\mathcal{R}}

\newtheorem{asm}{Assumption}
\newtheorem{theorem}{Theorem}
\newtheorem{lemma}{Lemma}
\newtheorem{definition}{Definition}

\theoremstyle{definition}
\newtheorem{remark}{Remark}

\newcommand{\V}{\overline{V}}
\newcommand{\Q}{\overline{Q}}
\newcommand{\Regret}{\operatorname{Regret}}
\newcommand{\overDelta}{\overline{\Delta}}

\title{Provably Efficient Action-Manipulation Attack Against Continuous Reinforcement Learning}

\author{%
        Zhi Luo$ ^{*1}$ 
        ~~~~~
        Xiyuan Yang$ ^{*1} $ 
        ~~~~~
        Pan Zhou$ ^{1}$
        ~~~~~
        Di Wang$ ^{2} $
        \vspace{0.5em}
        \\~
        $ ^1 $School of Cyber Science and Engineering, Huazhong University of Science and Technology\\
        ~~~~
        $ ^2 $King Abdullah University of Science and Technology \\
        ~~~~
        $ * $: Equal contribution
        \\
        \small
        \texttt{zhi\_luo@hust.edu.cn, xiyuan@hust.edu.cn,} \\
        \texttt{panzhou@hust.edu.cn, di.wang@kaust.edu.sa}
}

\begin{document}

\maketitle

\begin{abstract}
Manipulating the interaction trajectories between the intelligent agent and the environment can control the agent's training and behavior, exposing the potential vulnerabilities of reinforcement learning (RL). For example, in Cyber-Physical Systems (CPS) controlled by RL, the attacker can manipulate the actions of the adopted RL to other actions during the training phase, which will lead to bad consequences. Existing work has studied action-manipulation attacks in tabular settings, where the states and actions are discrete. As seen in many up-and-coming RL applications, such as autonomous driving, continuous action space is widely accepted, however, its action-manipulation attacks have not been thoroughly investigated yet. In this paper, we consider this crucial problem in both white-box and black-box scenarios. Specifically, utilizing the knowledge derived exclusively from trajectories, we propose a black-box attack algorithm named LCBT, which uses the Monte Carlo tree search method for efficient action searching and manipulation. Additionally, we demonstrate that for an agent whose dynamic regret is sub-linearly related to the total number of steps, LCBT can teach the agent to converge to target policies with only sublinear attack cost, i.e., $O\left(\mathcal{R}(T) + MH^3K^E\log (MT)\right)(0<E<1)$, where $H$ is the number of steps per episode, $K$ is the total number of episodes, $T=KH$ is the total number of steps, $M$ is the number of subspaces divided in the state space, and $\mathcal{R}(T)$ is the bound of the RL algorithm's regret. We conduct our proposed attack methods on three aggressive algorithms: DDPG, PPO, and TD3 in continuous settings, which show a promising attack performance. 
\end{abstract}

\section{Introduction}

Reinforcement learning (RL) aims to maximize the accumulated rewards through the interaction between the agent and the environment. With the development of RL, it has demonstrated outstanding performance in areas like robot control \citep{xu2020prediction, hayes2022practical}, autonomous system \citep{mattila2020inverse, kiran2021deep}, healthcare \citep{raghu2017continuous, yu2021reinforcement}, financial trading \citep{huang2018financial, avramelou2024deep} and etc. Nowadays, to accommodate more application scenarios, new algorithms such as DDPG \citep{lillicrap2015continuous} have been developed by researchers, extending the applicability of RL from discrete action space to continuous action space. However, the inherent vulnerability of RL might be exploited by attackers in this new setting \citep{ilahi2021challenges}. Therefore, studying RL's potential weaknesses is essential for constructing a secure RL system. 

Apart from tampering with observations \citep{foley2022execute, zhang2021robust, yang2020enhanced, pan2019characterizing, behzadan2017vulnerability, lin2017tactics}, rewards \citep{ma2019policy, zhang2020adaptive, ma2018data, huang2019deceptive, majadas2021disturbing, wu2022reward}, or the environment \citep{xu2021transferable, tanev2021adversarial, huang2017adversarial, boloor2019simple}, in this paper, we focus on attacks on the action space. For example, in Cyber-Physical Systems (CPS), during the training process, by tampering with the action signals output by the RL-based controller, the subsequent states and rewards are affected, thereby influencing the final learned control policies. However, the continuous action space contains infinitely many policies for decision making, attacks on which require different modeling and analysis methods. Therefore existing methods under tabular settings are not applicable \citep{liu2021provably, liu2024efficient}. Moreover, while there have been studies on action-manipulation attacks under continuous settings \citep{lee2020spatiotemporally,sun2020vulnerability,lee2021query}, they require prior knowledge such as the underlying MDP, the agent's models or the agent's algorithms, which is unrealistic in many scenarios. To this end, how to carry out efficient and practical action-manipulation attacks in continuous state and action spaces is more challenging and is still not well understood.

\begin{wrapfigure}{r}{0.5\linewidth}
\centering
\includegraphics[width=1\linewidth]{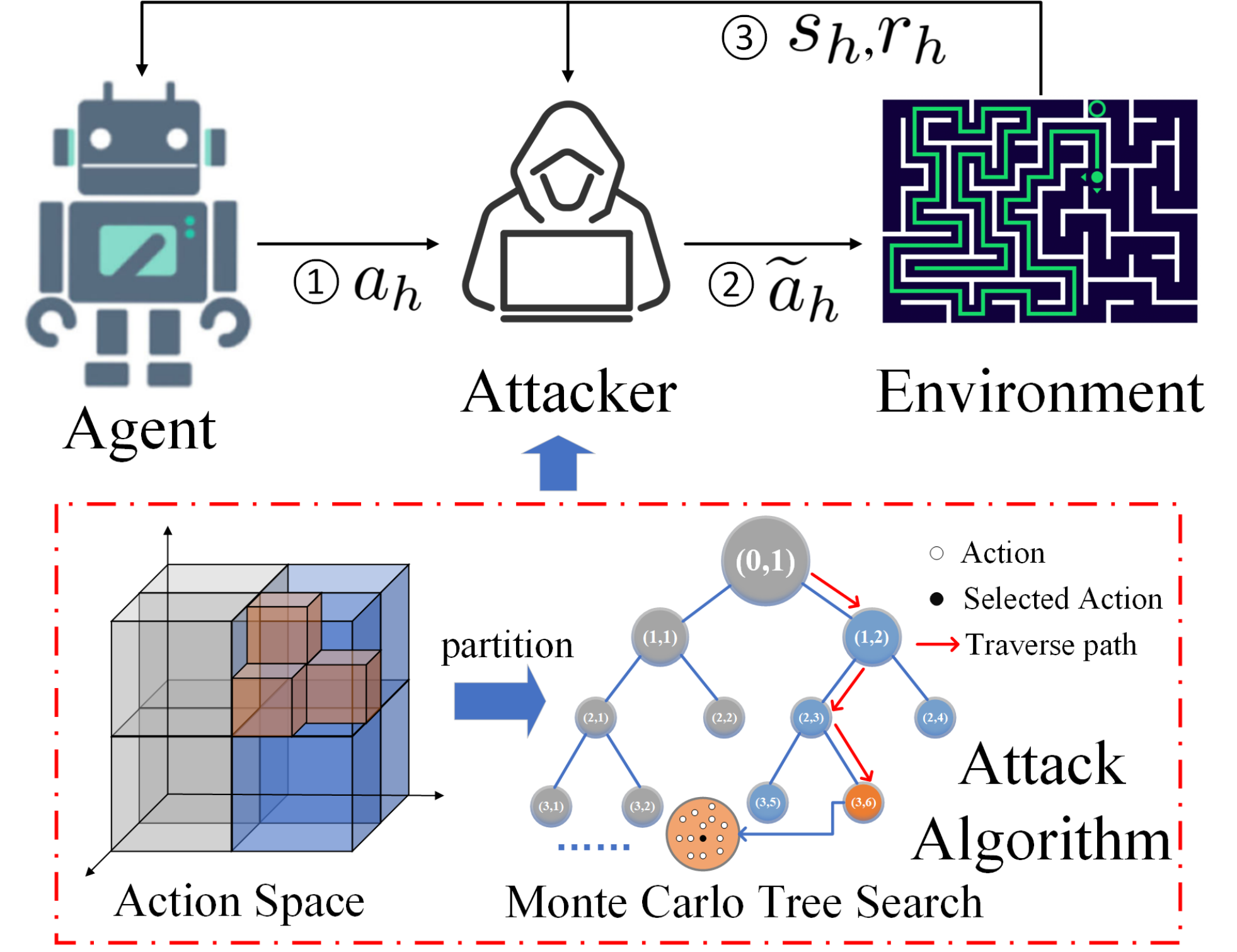}
\caption{Action-manipulation attack model. 
}
\label{attackmodel}
\vspace{-2mm}
\end{wrapfigure}

The key to achieving an efficient action-manipulation attack purpose lies in how to select appropriate actions from the action space to replace the original actions generated by the agent. 
In tabular RL, the attacker can evaluate the value of each action solely based on interaction trajectory information, thereby selecting the action that is most conducive to achieving the attack purpose. However, this method is not applicable when dealing with continuous values, as it involves modeling and partitioning of continuous action spaces. Thus, the following two questions are raised: \textbf{1). How to construct an efficient and practical attack method under continuous settings? 2). What is the efficiency of the attack?} The efficiency mentioned here includes runtime and sample complexity, i.e., both can be measured in continuous state and action spaces.

In this paper, we provide an affirmative answer by developing attack algorithms and establishing bounds on the attack cost for both white-box and black-box scenarios.  In the white-box attack scenario, the attacker possesses comprehensive knowledge of the involved Markov Decision Processes (MDPs), enabling them to more intuitively evaluate actions and design attack strategies. Conversely, in the black-box attack scenario, we propose a realistic attack without the knowledge of the underlying MDP, the agent's models and the agent's algorithms and solely rely on information extracted from the RL trajectory to devise attack strategies. In essentials, we use the Monte Carlo tree search method and incorporate it with the $Q$-values calculated from the trajectories to complete the search for actions as shown in Fig. \ref{attackmodel}. Our main contributions are as follows:
\begin{itemize}[leftmargin=*,nolistsep,nosep]
\item We have constructed a threat model for action-manipulation attacks under continuous state and action spaces, considering the attacker's knowledge, capability, and goal. To adapt to the continuous action space, we propose the concepts of the target action space and the target policy set. 
\item In the white-box scenario, based on our understanding of the underlying MDP, we intuitively propose the oracle attack method, and prove that the oracle attack can force the agent who runs a sub-linear-regret RL algorithm to choose actions according to the target policies with sublinear attack cost, i.e., $O\left(\R(T)\right)$. 
\item In the black-box scenario, we introduce an attack method called Lower Confidence Bound Tree (LCBT), with an attack cost bound of $O\left(\mathcal{R}(T) + MH^3K^E\log (MT)\right)$, where the second term represents the sublinear exploration cost of appropriate replacement actions. It can be demonstrated that the efficiency of this attack method can approach that of the oracle attack.
\item We employ the proposed attack methods to target three aggressive RL algorithms: DDPG \citep{lillicrap2015continuous}, PPO \citep{schulman2017proximal}, and TD3 \citep{fujimoto2018addressing}. The experimental results demonstrate the efficiency of our proposed methods. 
\end{itemize}


\section{Related Work}
\label{gen_inst}

The work \citep{liu2021provably} explored action-manipulation attacks against RL in a discrete action space and provided an attack cost bound, but it is not applicable in a continuous setting. \citep{lee2020spatiotemporally} and \citep{sun2020vulnerability} formalized the attack as an adversarial optimization problem, and used projected gradient descent (PGD) to manipulate actions. However, these two works require prior knowledge such as the model or specific algorithm used by the agent. The attacks in \citep{lee2021query} take place at test-time, where the trained agent and the environment are viewed as a new environment, and another adversarial DRL agent is trained to manipulate actions, but it can not change the policy of the agent itself and requires query permissions from the agent. For multi-agent RL, \citep{liu2024efficient} proposed an action-manipulation attack method under the white-box setting and provided an attack cost bound, but it is only applicable to tabular settings. Work \citep{mcmahan2024optimal} treats the attacker's problem as a MDP, that is, a new environment induced by the victim-attacker-environment interaction, and proposes a feasible attack framework, but does not clarify specific attack algorithms. Different from designing attack methods, the works \citep{tan2020robustifying} and \citep{tessler2019action} studied the defense methods under action-manipulation attacks. In comparison to the above works, we propose an action-manipulation attack algorithm in continuous settings that can control the policy learning of an agent and is free of the knowledge of the underlying MDP, the agent's models and the agent's algorithms. Table \ref{table:comparation} compares our work with other works on action-manipulation attacks against RL. The related work on tampering with observations, rewards, and the environment can be found in Appendix \ref{relatedwork}. 

\begin{table*}[!t]
\caption{Comparison of This Work with Other Action-manipulation Works. $S$ and $A$ respectively represent the number of states and actions in a discrete setting. $m$ is the number of agents in multi-agent RL.}
\label{table:comparation}
\centering
\resizebox{\textwidth}{!}{
\begin{tabular}{c|c|c|c|c|c}
\toprule[1.2pt]
\textbf{Method} & \textbf{Attacker's goal} & \textbf{\makecell{Continuous Setting \\ Support}} & \textbf{\makecell{The Knowledge \\ Demand of MDPs}} & \textbf{\makecell{The Knowledge Demand for \\ Agent's Models or Algorithms}} & \textbf{Attack Cost Bound} \\
\midrule
LCB-H \cite{liu2021provably} & Target Policy Learning & No &\XSolidBrush & \XSolidBrush & $O\left(H\R(T) + SAH^4\log (SAT)\right)$\\
\hline
$d$-portion \citep{liu2024efficient} & Target Policy Learning & No & \ding{52} & \XSolidBrush & $O\left(m^3\R(T)\right)$\\
\hline
LAS \citep{lee2020spatiotemporally} & Rewards Minimization & Yes & \ding{52} & \ding{52} & \textit{N/A}\\ 
\hline
VA2C-P \citep{sun2020vulnerability} & \makecell{Target Policy Learning/\\ Rewards Minimization} & Yes & \XSolidBrush & \ding{52} & \textit{N/A} \\
\hline
Query-based Attack \citep{lee2021query} & \makecell{Target State Reaching \\ (\textbf{test-time})} & Yes & \XSolidBrush & \XSolidBrush & \textit{N/A} \\
\hline
LCBT (\textbf{\emph{Ours}}) & Target Policy Learning & Yes & \XSolidBrush & \XSolidBrush & $O\left(\mathcal{R}(T) + MH^3K^E\log (MT)\right)$ \\
\bottomrule[1.2pt]
\end{tabular}
}
\vspace{-6mm}
\end{table*}

\section{Preliminaries and Problem Formulation}
\label{headings}

\textbf{Notations and Preliminaries:} This paper considers a finite-horizon MDP over continuous domains. Such an MDP can be defined as a 6-tuple $\mathcal{M} = \left( {S,A,H,P,R,\mu } \right)$, where $S$ and $A$ are respectively bounded continuous state and action spaces, $H$ denotes the number of steps per episode, $P$ is the transition kernel, $R_h:S \times A \rightarrow [0,1]$ represents the reward function at step $h$ and $\mu$ is the initial state distribution. Define $K$ as the total number of episodes and $T=KH$ as the total number of steps. For each episode $k \in [K]$, a trajectory $\gamma \sim \pi$ generated by policy $\pi$ is a sequence $\left\{ {{s_1},{a_1},{r_1},{s_2},{a_2},{r_2},...,{s_H},{a_H},{r_H},{s_{H+1}}} \right\}$. A policy $\pi$ can be evaluated by the expected rewards. Formally, we define $V_h^\pi (s) = {\mathbb{E}}\left[ \sum_{{h^\prime } = h}^H r_{h^\prime} |{s_h} = s, \pi\right]$ and $Q_h^\pi \left( {s, a} \right) = \mathbb{E}\left[\sum_{{h^\prime } = h}^H r_{h^\prime}|{s_h} = s, {a_h} = a, \pi \right]$. For notation simplicity, we denote $V_{H+1}^\pi=0$, $Q_{H+1}^\pi=0$ and $P_hV_{h+1}^{\pi}(s,a) = \mathbb{E}_{s'\sim P_h(\cdot|s,a)}\left[V_{h+1}^{\pi}(s')\right]$. A deterministic policy is a policy that maps each state to a particular action. For a deterministic policy $\pi$, we use $\pi_h(s)$ to denote the action $a$ which satisfies $\pi_h(a|s)=1$. 

We assume that $H$ is finite, and because $r_h\in[0,1]$, the $Q$-values and $V$-values are bounded. The purpose of RL is to find the optimal policy $\pi^*$, which gives the optimal value function $V_h^*\left( s \right) = V_h^{\pi^*}\left( s \right) = \sup_{\pi}V_h^\pi(s)$ and $Q_h^*(s,a) = Q_h^{\pi^*}(s,a) = \sup_{\pi} Q_h^{\pi}(s,a)$ for all $s \in S$, $a \in A$ and $h \in [H]$. To assess an RL algorithm's performance over $K$ episodes, we quantify a metric known as regret, which is defined as: 
${\mathop{\rm Regret}\nolimits} (K) = \sum_{k = 1}^K {\left[ {V_1^*\left( {s_1^k} \right) - V_1^{{\pi^k}}\left( {s_1^k} \right)} \right]}$.


\textbf{Threat Model Description:} The attacker aims for the agent to learn a deterministic target policy $\pi^{\dagger}$. However, in a continuous action space that contains infinitely many policies, it is difficult for the agent to learn a specific policy exactly. In this study, the attacker expects the agent to learn a policy that approaches the target policy. The attacker measures the similarity of policies based on the degree of closeness of actions produced in the same state. Therefore, the attacker sets a dissimilarity radius of $r_a$, which requires that the dissimilarity between the actions produced by the agent's learned policy and the target policy does not exceed $r_a$ w.r.t a distance function $l$ employed by the attacker (e.g. Euclidean distance). Define $\mathcal{A}^{\dagger}_h(s) = \{a: l(a,\pi^{\dagger}_h(s))\leq r_a\}$ as the target action space for state $s$. Intuitively, $\mathcal{A}^{\dagger}_h(s)$ is the set of acceptable malicious actions for the adversary. According to the target action space, we define the target policy set as $\Pi^{\dagger} = \{\pi:\pi_h(s) \in \mathcal{A}^{\dagger}_h(s), \forall s \in S, h \in [H]\}$, which represents the set of acceptable malicious policies. 
We describe the threat model in this paper from the perspective of attacker's knowledge, capability, and goal. 

\begin{itemize}[leftmargin=*,nolistsep,nosep]
\item \textbf{Attacker's Knowledge:} In the black-box scenario, the attacker has no prior knowledge about the underlying MDP, the agent's models or the agent's algorithms. It only has access to the interaction information between the agent and the environment, i.e., $s^k_h$, $a^k_h$, and $r^k_h$. 

\item \textbf{Attacker's Capability:} In each step $h \in [H]$ at episode $k \in [K]$, 
the attacker can manipulate the action $a^k_h$ generated by the agent and change it to another action $\widetilde{a}^k_h$. If the attacker decides not to attack, $\widetilde{a}^k_h = a^k_h$. 

\item \textbf{Attacker's Goal:} 
The attacker's goal is to manipulate the agent into picking its actions according to the policies within $\Pi^{\dagger}$, while simultaneously minimizing the attack cost. Specifically, let $\tau$ be the step set whose element is the step when the attacker launches attack, i.e., $\tau=\{(k,h):\widetilde{a}^k_h \neq a^k_h, k \in [1, K], h \in [1, H]\}$. And let $\alpha$ be the step set whose element is the step when the action $a^k_h$ generated by the agent meets the condition $a^k_h \notin \mathcal{A}^{\dagger}_h(s^k_h)$, i.e., $\alpha=\{(k,h):a^k_h \notin \mathcal{A}^{\dagger}_h(s^k_h),k \in [1,K], h \in [1,H]\}$. The attacker aims to minimize both the cost $|\tau|$ and the loss $|\alpha|$. 
\end{itemize}

\textbf{Attack Scenario Analysis and Additional Notations:} 
We represent the optimal policy in the policy set $\Pi^{\dagger}$ as $\pi^o \in \Pi^{\dagger}$, which gives the value $V^{\pi^o}_h(s) = \sup_{\pi\in \Pi^{\dagger}}V^{\pi}_h(s),\forall s \in S, h \in [1,H]$. For notation simplicity, we denote $V^{\dagger}_h(s):=V^{\pi^{\dagger}}_h(s)$ and $V^o_h(s):=V^{\pi^o}_h(s)$. If the attacker can make the agent believe that the policy $\pi^o$ is the globally optimal policy, then he can induce the agent to learn the policies in $\Pi^{\dagger}$. For the feasibility of the attack, $\pi^o$ must not be the worst global policy, here we require that $\pi^o$ satisfies:
\begin{equation}
\label{conditionmin}
    V^o_h(s) > Q^o_h(s,a^-_h(s)), \forall s \in S, h \in [H], 
\end{equation}
where $a^-_h(s) := \mathop{\arg\min}_{a \in A}Q^o_h(s,a)$. Define the minimum gap $\Delta_{min}$ by $\Delta_{min} = \min_{h\in [H], s\in S}\left(V^o_h(s) - Q^o_h(s,a^-_h(s))\right)$. Given (\ref{conditionmin}), $\Delta_{min} > 0$ will hold. Therefore, the attacker should try not to choose the worst global policy as the target policy, otherwise, there may potentially exist $V^o_h(s)=V^{\dagger}_h(s)=\inf_{\pi}V^{\pi}_h(s)$. A summary of the notations can be found in Appendix \ref{notationtable}.

\section{Attack Strategy and Analysis}

\subsection{White-box Attack}

In the white-box setting, the attacker has comprehensive knowledge of the MDP $\mathcal{M}$. Therefore, the attacker can compute the $Q$ and $V$-values of any policy, and the action $a^-_h(s)$ is known to the attacker. 


In order to make the agent learn the policies in $ \Pi^{\dagger}$, the attacker can mislead the agent into believing $\pi^o$ is the optimal policy. We now introduce an effective oracle attack strategy. Specifically, at the step $h$, if the action selected by the agent is within the target action space, i.e., $a_h \in \mathcal{A}^{\dagger}_h(s_h)$, the attacker does not launch an attack, i.e., $\widetilde{a}_h = a_h$. Otherwise, the attacker launches an attack and sets $\widetilde{a}_h = a^-_h(s_h)$ as the worst action. In earlier $\K$ episodes, the attacker does not carry out any attacks, and $\K$ can be set to 0, or a number of order $o(K)$. Under non-attack conditions, there is continuity between the $Q$-values of actions in $\mathcal{A}^{\dagger}_h(s_h)$ and those of other actions (the oracle attack can greatly disrupt this continuity). Therefore, an appropriate $\K$ value can enhance the hit rate of the RL algorithm for the target action space during the initial exploration phase, which is beneficial for learning the policies within $\Pi^{\dagger}$ in the later stages. This becomes particularly evident when the proportion of the target action space $\mathcal{A}^{\dagger}_h(s_h)$ in the total action space $A$ is quite small. 
The detailed procedure of the oracle attack algorithm is delineated in Appendix \ref{oraclealg}. 
Then, we have: 
\begin{lemma}
\label{lem:pre}
    If $\Delta_{min} > 0$ and the attacker follows the oracle attack scheme, then from the agent's perspective, $\pi^o$ is the optimal policy.
\end{lemma}
The detailed proof is included in Appendix \ref{prooflemma1}. With the condition $\Delta_{min} > 0$, the upper bounds of $|\tau|$ and $|\alpha|$ under the oracle attack can be obtained. The detailed proof is included in Appendix \ref{prooftheorem1}.
\begin{theorem}
\label{theorem:oracle}
    In the white-box setting, with a probability at least $1-\delta_2$, the oracle attack will force the agent to learn the  policies in $\Pi^{\dagger}$ with $|\tau|$ and $|\alpha|$ bounded by
    \begin{align}
        |\tau| &\leq \frac{\operatorname{Regret}(K) + 2H^2 \sqrt{\ln(1/\delta_2) \cdot \operatorname{Regret}(K)}}{\Delta_{min}}, 
    \end{align}
    and $|\alpha| \leq |\tau| + H\K$. 
\end{theorem}
\begin{remark}
    Theorem \ref{theorem:oracle} indicates that under the oracle attack, the attack cost depends on the performance of the RL algorithm used by the agent itself. 
    \textcolor{black}{Moreover, if $r_a$ is larger, the range of actions covered by the policies in $\Pi^{\dagger}$ is larger. Under the same target policy, the value of the corresponding $\pi^o$ might be higher, which would possibly yield a larger $\Delta_{min}$. Also, a larger target action space implies a higher hit rate of the RL algorithm towards it.}
\end{remark}



\begin{algorithm}[!t] 
 	\caption{The LCBT attack algorithm.}   	\label{alg:Framwork} 
 	\begin{algorithmic}[1]
 	    \REQUIRE ~~\\ 
 		Target policy $\pi^\dag$.
 	    \STATE Initialize $\T^h_{1}=\{(0,1),(1,1),(1,2)\}$, $\hat{Q}^h_{1,1}(1) = \hat{Q}^h_{1,2}(1) = 0$, $L^h_{1,1}(1)=L^h_{1,2}(1)= - \infty$, and $T^h_{1,1}(1)=T^h_{1,2}(1)=0$, for all $h \in [1,H]$.  
 		\FOR{episode $k = 1, 2, \dots, K$}
 		\STATE Receive $s_1^k$. Initialize the set of trajectory $traj = \{s_1^k\}$.
 		\FOR{step $h = 1, 2, \dots, H$}
 		\STATE The agent chooses an action $a_h^k$.
 		\IF {$k \leq \K$ or $a_h^k \in \mathcal{A}^{\dagger}_h(s_h^k)$}
 		\STATE The attacker does not attack, i.e., $\widetilde{a}_h^k =a_h^k$, \textcolor{black}{and sets $w_h = 1$}.
 		\ELSE 
            \STATE Take $\{(D^h_{k},I^h_{k}), P^h_{k}\}\leftarrow \text{WorTraverse}(\T^h_k)$, and set $\widetilde{a}^k_h=a_{D^h_k,I^h_k}$, $w_h = 0$. 
 		\ENDIF
 		\STATE The environment receives action $\widetilde{a}_h^k$, and returns the reward $r_h^k$ and the next state $s_{h+1}^k$.
 		\STATE Update the trajectory by plugging $\widetilde{a}_h^k$, $r_h^k$ and $s_{h+1}^k$ into $traj$.
        \ENDFOR
        \STATE Set the cumulative reward $G_{H+1:H+1}=0$ and the importance ratio $\rho_{H+1:H+1}=1$.
 	\FOR{step $h = H, H-1, \dots, 1$ 
 		} 

            \IF{$k \leq \K$ or $a_h^k \in \mathcal{A}^{\dagger}_h(s^k_h)$}
            \STATE Continue. 
            \ENDIF
            
 		\STATE
   $t^h_k=T^h_{D^h_{k},I^h_{k}}(k)\leftarrow T^h_{D^h_{k},I^h_{k}}(k)+1$.
            \STATE Use Eq.(\ref{calQ}) to update the value $\hat{Q}^h_{D^h_{k},I^h_{k}}(s^k_h, a_{D^h_{k},I^h_{k}})$. 
            \STATE $G^k_{h:H+1}=r_h^k+G^k_{h+1:H+1}$, $\rho^k_{h:H+1}=w_h \cdot \rho^k_{h+1:H+1}$. 
            \IF{$\nu_1\rho^{D^h_k} \geq \frac{H-h+1}{\sqrt{2t^h_k}}\sqrt{\ln \Big(\frac{2Mk\sum_{h=1}^H|\T^h_k|}{\delta_1}}\Big)$ AND $(D^h_{k},I^h_{k}) \in \operatorname{leaf}(\T^h_k)$}
            \STATE $\T^h_k \leftarrow \T^h_k \cup \{(D^h_k+1,2I^h_k-1),(D^h_k+1,2I^h_k)\}$. 
            \STATE $L_{D^h_k+1,2I^h_k-1}(k)=L_{D^h_k+1,2I^h_k}(k)=-\infty$. 
 		\ENDIF
            \FORALL{$(D,I) \in \T^h_k$}
            \STATE Use Eq.(\ref{calL}) to update the value $L^h_{D,I}(k)$.
            \STATE Use Eq.(\ref{eq:def.stats.upB}) to update the value $B^h_{D,I}(k)$ backward from leaf nodes.
            \ENDFOR
            \ENDFOR
 		\ENDFOR
 	\end{algorithmic}
\label{lcbt}
\end{algorithm}

\subsection{Black-box Attack}

In the black-box attack scenario, the attacker has no knowledge of the underlying MDP process and the algorithm used by the agent but only knows the interaction information: $s^k_h$, $a^k_h$, and $r^k_h$. Without the knowledge of $a^-_h(s)$, the attacker can not determine $\widetilde{a}$ and fails to attack.

To counteract this issue, in our proposed attack, the attacker can employ a technique that methodically identifies and approximates $a^-_h(s)$. In particular, an infinite $Q$-value tree structure is developed to encapsulate the action space $A$, facilitating cluster-level analysis thereby reducing the scale of actions being analyzed. Meanwhile, we use important sampling to calculate the $Q$-values of $\pi^o$ and use Hoeffding’s inequality to establish the LCB of the $Q$-values. The attack algorithm has been designated as \textit{LCBT} (Lower Confidence Bound Tree), in relation to its utilization of a tree structure. Before introducing the technical details of the LCBT algorithm, we define the notion of dissimilarity functions.

\begin{algorithm}[!t]
\begin{algorithmic}[1]
\REQUIRE ~~\\
$\T^h_k$
\STATE $(D,I) \leftarrow (0,1)$, $P\leftarrow (0,1)$ 
\WHILE {$(D,I)\notin \operatorname{leaf}(\T^h_k)$}
\IF{$B^h_{D+1,2I-1} \leq  B^h_{D+1,2I}$} 
\STATE $(D,I)\leftarrow(D+1,2I-1)$
\ELSE
\STATE $(D,I)\leftarrow (D+1,2I)$
\ENDIF
\STATE $P\leftarrow P \cup \{(D,I)\}$
\ENDWHILE
\RETURN $(D,I)$ and $P$
\end{algorithmic}
\par
\caption{The \textit{WorTraverse} function.}
\label{wortraverse}
\par
\end{algorithm}

\begin{definition}
\label{metric}
    The state space $S$ is equipped with a dissimilarity function $l_s : S^2 \rightarrow \mathbb{R}$ such that $l_s(s,s') \geq 0$ for all $(s,s') \in S^2$ and $l_s(s,s) = 0$. Likewise, The action space $A$ is equipped with a dissimilarity function $l_a : A^2 \rightarrow \mathbb{R}$ such that $l_a(a,a') \geq 0$ for all $(a,a') \in A^2$ and $l_a(a,a) = 0$. 
\end{definition}
For a subset $D \subseteq S$, the diameter of it is defined as $diam_s(D) := \sup_{x,y \in D} l_s(x,y)$. Likewise, we define $diam_a(P) := \sup_{x,y \in P} l_a(x,y)$ where $P \subseteq A$. It should be noted that the distance function $l$ used in the target action space $\mathcal{A}$ can be different from $l_a$. 

\textbf{Action Cover Tree:}
As the action space $A$ is continuous, we use a binary tree $\T$ to discretize the action space and reduce the possible options for selection. In the cover tree, we denote by $(D,I)$ the node at depth $D \geq 0$ with index $I \in [1,2^D]$ among the nodes at the same depth. Clearly, the root node is $(0,1)$. The two children nodes of $(D,I)$ are denoted by $(D+1,2I-1)$ and $(D+1,2I)$ respectively. For each node $(D,I)$, a continuous subset $\mathcal{P}_{D,I} \subseteq A$ of actions is divided from the action space and associated with this node. The $\mathcal{P}_{D,I}$ can be determined recursively as $\mathcal{P}_{0,1} = A$, $\mathcal{P}_{D,I}=\mathcal{P}_{D+1,2I-1} \cup \mathcal{P}_{D+1,2I}$, and $\mathcal{P}_{D,I} \cap \mathcal{P}_{D,J} = \emptyset, \forall I, J \in [1, 2^D]$. 
For each node $(D,I)$, an action $a_{D,I}$ is selected from $\mathcal{P}_{D,I}$ to represent the node. Whenever $(D,I)$ is sampled, action $a_{D,I}$ is selected. Until the beginning of episode $k$, for each step $h \in [H]$, an action coverage tree is maintained and it is denoted as $\T^h_k$, and $|\T^h_k|$ is the node number. 


\textbf{State Partition:}
In order to better evaluate $Q$-values, we also need to discretize the continuous state space. Here, we partition it into $M$ subspaces $S_1, S_2,...,S_{M-1},S_M$ which satisfy the following conditions: $S = \mathop{\cup}_{m=1}^{M} S_m$ , and $S_i \cap S_j = \emptyset$ for $\forall i,j \in [1,M]$.
Define $i(s):s \in S_{i(s)}$, which represents the number of the subspace to which state $s$ belongs. 



We introduce constants $\nu_1$, $0<\rho<1$, which satisfy $diam_a(\mathcal{P}_{D,I}) \leq \nu_1 \rho^D$, as well as constants $L_s$ and $d_s$, which satisfy $L_sd_s < \Delta_{min}/2$ and $diam_s(S_m) \leq L_sd_s$. The related assumptions can be seen in Appendix \ref{lcbtasm}.





\textbf{LCB Calculation:} LCB is the pessimistic estimate of unknown $Q^o_h(s,a)$. In the black-box attack, at the step $h$ in episode $k$, the attacker will select a node $(D,I)$ and set $\widetilde{a}^k_h$ as $a_{D,I}$ when $a^k_h \notin \mathcal{A}^{\dagger}_h(s^k_h)$. In the LCBT attack algorithm, for each node $(D,I)$ in $\T^h_k$, $\hat{Q}^h_{D,I}(s^k_h,a_{D,I})$ is used to evaluate $Q$-values, which is calculated by 
\begin{equation}
\label{calQ} 
\hat{Q}^h_{D,I}(s^k_h,a_{D,I}) = (1-\frac{1}{T^h_{D,I}(k)})\hat{Q}^h_{D,I}(s^{\gamma^h_{D,I}(k)}_h,a_{D,I}) + \frac{1}{T^h_{D,I}(k)} (r_h^k + G^k_{h+1:H+1} \cdot \rho^k_{h+1:H+1}),
\end{equation}
where $T^h_{D,I}(k) = |\phi^h_{D,I}(k)|$, and $\phi^h_{D,I}(k) = \{ \gamma:\K < \gamma < k, s^{\gamma}_h \in S_{i(s^k_h)}, a^{\gamma}_h=a_{D,I}\}$, is defined as the set of episodes in which the current state belonged to subinterval $S_{i(s^k_h)}$ and node $(D,I)$ was selected by the attacker at the step $h$ until the beginning of the episode $k$. $\gamma^h_{D,I}(k) = \max\{\gamma: \gamma \in \phi^h_{D,I}(k)\}$ represents the latest episode before $k$ in which the current state belonged to the subinterval $S_{i(s^k_h)}$ and node $(D,I)$ was selected in the step $h$. Let $(D^h_{k},I^h_{k})$ be the node selected by the attacker at the step $h$ in episode $k$ in $\T^h_k$. $G^k_{h:H} = \sum_{h'=h}^H r^k_h$ is the cumulative reward. The importance sampling ratio is calculated by $\rho^k_{h:H}=\prod_{h'=h}^H \frac{\mathbb{P}(\widetilde{a}^k_{h'}|s^k_{h'}, \pi^o)}{\mathbb{P}(\widetilde{a}^k_{h'}|s^k_{h'}, b^k_{h'})}$, where 
$b^k$ is the behavior policy that generates trajectory $\{s^k_1, \widetilde{a}^k_1, r^k_1, s^k_2, \widetilde{a}^k_2, r^k_2,... , s^k_H, \widetilde{a}^k_H, r^k_H, s^k_{H+1}\}$. $\mathbb{P}(\widetilde{a}^k_h|s^k_h,b^k_h)$ is
\begin{equation}
\label{chia}
\mathbb{P}(\widetilde{a}^k_h|s^k_h,b^k_h)\!=\!
\begin{cases}
1 \quad\quad\quad \text{if } \widetilde{a}^k_h = a^k_h \text{ and } a^k_h \in \mathcal{A}^{\dagger}_h(s^k_h),
\\
&
\\
1 \quad\quad\quad \text{if } \widetilde{a}^k_h = 
a_{D^h_k,I^h_k} \text{ and } a^k_h \notin \mathcal{A}^{\dagger}_h(s^k_h),
\end{cases}
\end{equation}
otherwise, the value is $0$. We employ the action trajectories in the target action space to assess $\pi^o$, hence, we set $\mathbb{P}(\widetilde{a}_h|s_h, \pi^o_h) = \mathbb{I}\{\widetilde{a} \in \mathcal{A}^{\dagger}_h(s)\}$. For the indicator function $\mathbb{I}\{\xi\}$, if event $\xi$ is established $\mathbb{I}\{\xi\}=1$, otherwise $\mathbb{I}\{\xi\}=0$. We set $\rho^k_{H+1:H+1}=1$, $G^k_{H+1:H+1}=0$ and $\rho^k_{h:H+1}=\rho^k_{h:H}$, $G^k_{h:H+1}=G^k_{h:H}$. According to the definition of $b^k$, we have $V^{b^k}_h(s)=\mathbb{E}[G^k_{h:H}|s^k_h=s]$ and $V^o_h(s)=\mathbb{E}[\rho^k_{h:H}G^k_{h:H}|s^k_h=s]$. 

The states used to calculate the $\hat{Q}^h_{D,I}(s, a_{D,I})$ value are limited to a subinterval, instead of being fixed. Also, each tree node $(D,I)$ covers the action space $\mathcal{P}_{D,I}$ , despite being represented by action $a_{D,I}$. Thus, the lower confidence bound of node $(D,I)$ can be calculated by 
\begin{equation}
\label{calL}
L^h_{D,I}(k) = \hat{Q}^h_{D,I}(s^k_h,a_{D,I}) - \frac{H-h+1}{\sqrt{2T^h_{D,I}(k)}}\sqrt{\ln \left(\frac{2Mk\sum_{h=1}^H|\T^h_k|}{\delta_1}\right)} - L_sd_s - \nu_1\rho^D. 
\end{equation}
We use coefficients $L_sd_s$ and $\nu_1\rho^D$ to measure the level of uncertainty generated by the state subinterval and the node's action space, respectively. The second term denotes the radius of the confidence interval, which is derived utilizing Hoeffding's inequality for the estimation of $Q$-values from $\hat{Q}^h_{D,I}(s, a_{D,I})$.

\textbf{Worst Node Selection:} 
When in the step $h$ of episode $k$, if the action $a^k_h$ chosen by the agent is not in $\mathcal{A}^{\dagger}_h(s^k_h)$, then the attacker will utilize the $L$-values to choose a node $(D,I)$, and use $a_{D,I}$ to replace $a^k_h$, that is, set $\widetilde{a}^k_h$ to $a_{D,I}$ (\textbf{\textcolor{black}{Line 4-13}}). Due to $\nu_1\rho^{D+1} < \nu_1\rho^D$, the child node has a smaller structural resolution, which reduces the uncertainty in the LCB estimate. Define the $B$-values:
\begin{equation}
\label{eq:def.stats.upB}
B^h_{D,I}(k)\!=\!
\begin{cases}
L^h_{D,I}(k) \quad\quad\quad\quad\quad\quad\quad\quad\text{if}\,\,\,(D,I)\!\in\!\operatorname{leaf}(\T^h_k),
\\
&
\\
\begin{aligned}
\max\Big[ L^h_{D,I}(k), \!\!\!\min_{j \in \{2I-1,2I\}} \!\!\!B^h_{D+1,j}(k)\Big]
 \end{aligned} \text{\quad otherwise}.
\end{cases}
\end{equation}
The $B$-values are designed to have a tighter lower bound on $Q^o_h(s^k_h, a_{D,I})$ by taking the maximum between $L^h_{D,I}(k)$ for the current node, and the minimum lower bound of the node's two child nodes. Based on the $B$-values, the attacker traverses the tree $\T^h_k$ from the root node with smaller $B$-values to the leaf node, and the path is represented as $P^h_k$. The traverse function is shown in Algorithm \ref{wortraverse}.  

\textbf{Update and Node Expansion:}
After one episode, the attacker begins to utilize the interaction data between the agent and the environment to update the $\hat{Q}$, $L$, and $B$ values of nodes (\textbf{\textcolor{black}{Line 15-30}}). A critical step is deciding when to expand a node into its two child nodes to reduce the uncertainty that is caused by the size of the node. Intuitively, a node should be expanded when it is chosen a particular number of times such that the radius of the confidence interval is approximately equal to the node's size. This occurs when the uncertainty brought by the node's size begins to dominate. Therefore, for node $(D^h_k,I^h_k)$, the attacker expands it into its two child nodes when 
    $\nu_1\rho^{D^h_k} \geq \frac{H-h+1}{\sqrt{2t^h_k}}\sqrt{\ln \left(\frac{2Mk\sum_{h=1}^H|\T^h_k|}{\delta_1}\right)}$ 
is true, and set the $L$-values of the two child nodes as $-\infty$. Because the \textit{WorTraverse} function selects nodes with smaller $B$-values, nodes containing $a^-_h(s^k_h)$ are more likely to be expanded, thereby reducing the uncertainty caused by node size and sufficiently approximating action $a^-_h(s^k_h)$.

In earlier $\K$ episodes, the attacker does not carry out any attacks, and $\K$ can be set to 0, or a number of order $o(K)$.



Before presenting the main theorem, it is important to note that if the attacker and the environment are considered as a single entity, the resulting environment is non-stationary, i.e., the reward functions and probability transition function may change over episodes. 
As a result, the measurement of regret also undergoes changes. In plain terms, dynamic regret measures the degree of regret resulting from comparing the policy adopted by an agent with the optimal policy of each specific episode in hindsight. On the other hand, static regret only compares the agent's policy with the optimal fixed policy derived from combining all episodes. \textcolor{black}{According to \citep{fei2020dynamic},} the definition of the dynamic regret is $\operatorname{D-Regret}(K):=\sum_{k \in[K]}\left[V_1^{\pi^{*, k}, k}\left(s_1^k\right)-V_1^{\pi^k, k}\left(s_1^k\right)\right]$, where $\pi^{*, k}=\operatorname{sup}_{\pi} V_1^{\pi, k}\left(s_1^k\right)$ is the optimal policy of episode $k$.

Our main theorem is the upper bound of cost $|\tau|$ and loss $|\alpha|$ in the context of using the LCBT attack algorithm. The detailed proof is included in Appendix \ref{prooftheorem2} and \ref{nodenumber}. The two fundamental lemmas can be seen in Appendix \ref{prooflemma2} and \ref{prooflemma3}. 
\begin{theorem}
\label{theorem:lcbt}
    In the black-box setting, with a probability at least $1-\delta_1-\delta_2$, the LCBT attack will force the agent to learn the  policies in $\Pi^{\dagger}$ with $|\tau|$ and $|\alpha|$ bounded by
    \begin{equation}
        |\tau| \leq \frac{2\left(\operatorname{D-Regret}(K) + 2H^2 \sqrt{\ln(\frac{1}{\delta_2}) \cdot \operatorname{D-Regret}(K)}\right)}{\Delta_{min} - 2 L_sd_s} + \frac{18 M H^2 \ln(\frac{2MHK^2}{\delta_1}) \sum_{h=1}^H |\T^h_K|}{(\Delta_{min} - 2 L_sd_s)^2}, 
    \end{equation}
    and $|\alpha| \leq |\tau| + H\K$. Where $|\T^h_K| \leq O(K^E), E = \log_{2\rho^{-2}}2 < 1$ for $\forall h \in [H]$.
\end{theorem}
\begin{remark}
When $\operatorname{D-Regret}(K) \leq O\left(\frac{MH^3K^E\ln(\frac{2MHK^2}{\delta_1})}{\Delta_{min} - 2 L_sd_s}\right)$, we have the upper bound $O\left(\frac{MH^3K^E\ln(\frac{2MHK^2}{\delta_1})}{(\Delta_{min} - 2 L_sd_s)^2}\right)$. In other words, a lower attack cost may be incurred when the RL algorithm utilized by the agent works better in non-stationary environments. Compared with the results of the oracle attack (Theorem \ref{theorem:oracle}), the additional second term can be seen as the cost of exploring the worst action. \textcolor{black}{The denominator $\Delta_{min} - 2 L_sd_s$ and the $M$ in the numerator indicate that an appropriate fine-grained division of the state space is conducive to the implementation of the LCBT attack.} It should be noted that we use Hoeffding's inequality to calculate the LCB and the attack cost can be potentially improved when using Bernstein-type concentration inequalities. 
\end{remark}

The time complexity of the LCBT algorithm is $O(H K^{1+E}+H K \cdot \log_{\rho^{-2}}K)$, and space complexity is $O(MHK^E)$. The proof can be found in Appendix \ref{timespace}. Moreover, we calculated the percentage of time spent executing the LCBT algorithm during the training phase out of the total time.

\section{Numerical Results}
\label{expresults}

In this section, we conduct the oracle attack and the LCBT attack on three aggressive continuous RL algorithms, including DDPG, PPO, and TD3, in three different environments. The experiments are run on a machine with NVIDIA Quadro RTX 5000. 
Environment $1$ involves a one-dimensional continuous control problem with $s \in \left[-1,1\right]$ and $a \in \left[-1,1\right]$ where a slider moves on a rail, receiving positive rewards proportional to the move distance, with a negative reward when it falls off. Environment $2$ describes a two-dimensional continuous control problem with $\textit{\textbf{s}} \in \left[0,8\right]^2$ and $\textit{\textbf{a}} \in \left[-1,1\right]^2$ where a vehicle moves on a two-dimensional plane, receiving rewards \textcolor{black}{negatively proportional} to the distance from the central point for every step. 
Environment $3$ is a five-dimensional version of Environment $2$ with $\textit{\textbf{s}} \in \left[0,8\right]^5$ and $\textit{\textbf{a}} \in \left[-1,1\right]^5$. A detailed description of the environments can be found in Appendix \ref{expset}. The target policy $\pi^{\dagger}$ is trained by constraining the movement range of both the slider and vehicle. 
In Appendix \ref{additionalexp}, we present the experimental results of Environment 3.

\begin{figure*}[!t]
\centering
\subfigure[Attack DDPG]{
\includegraphics[width=0.32\columnwidth]{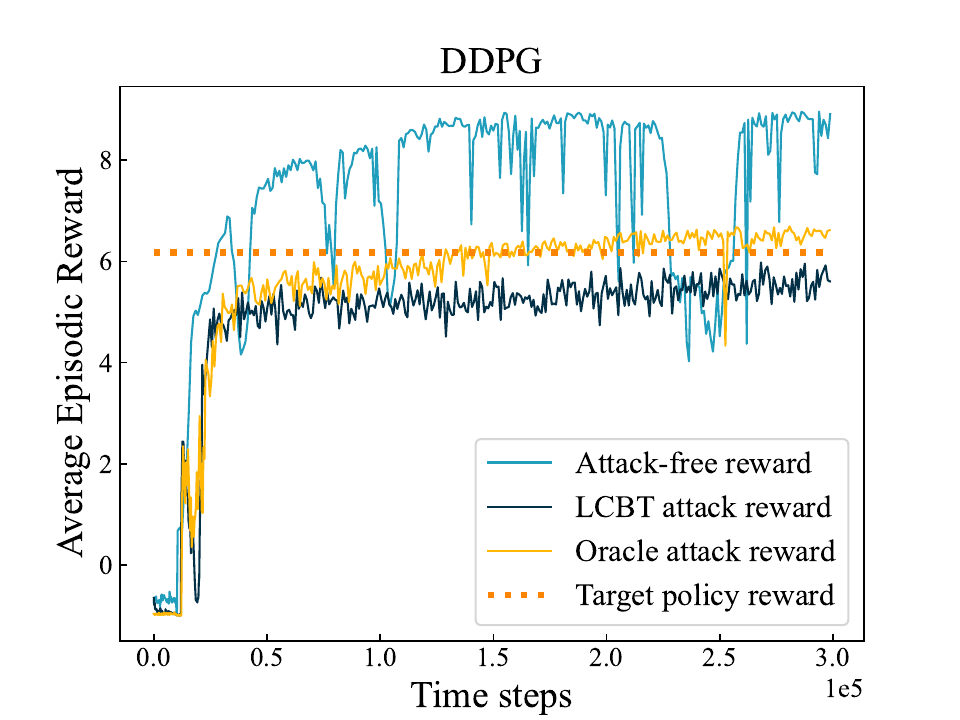}}
\hfill
\subfigure[Attack PPO]{
\includegraphics[width=0.32\columnwidth]{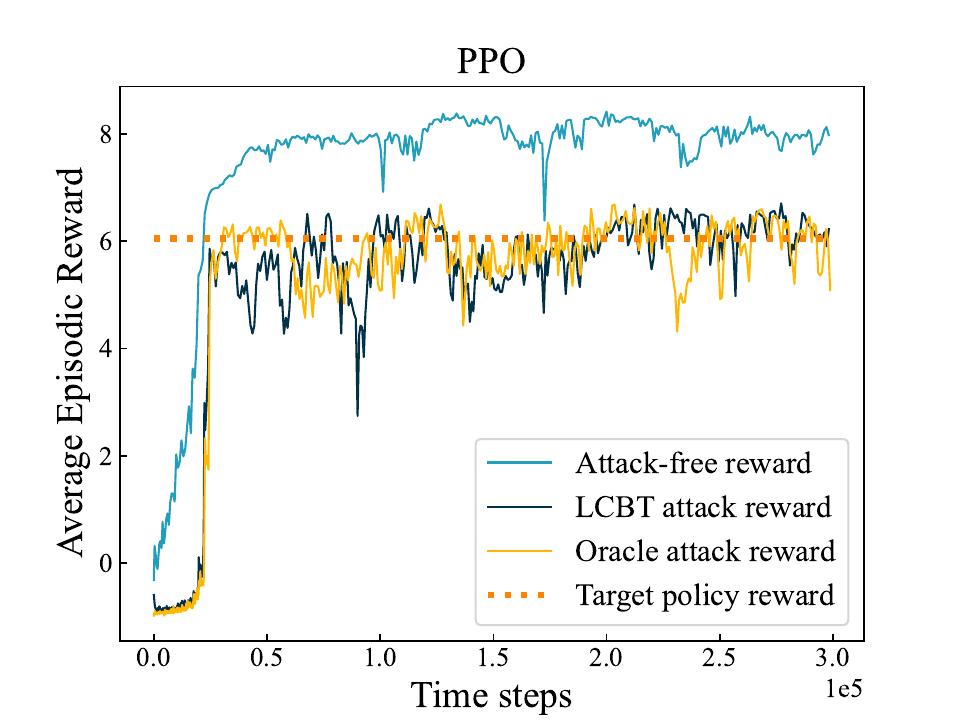}}
\hfill
\subfigure[Attack Cost]{
\includegraphics[width=0.32\columnwidth]{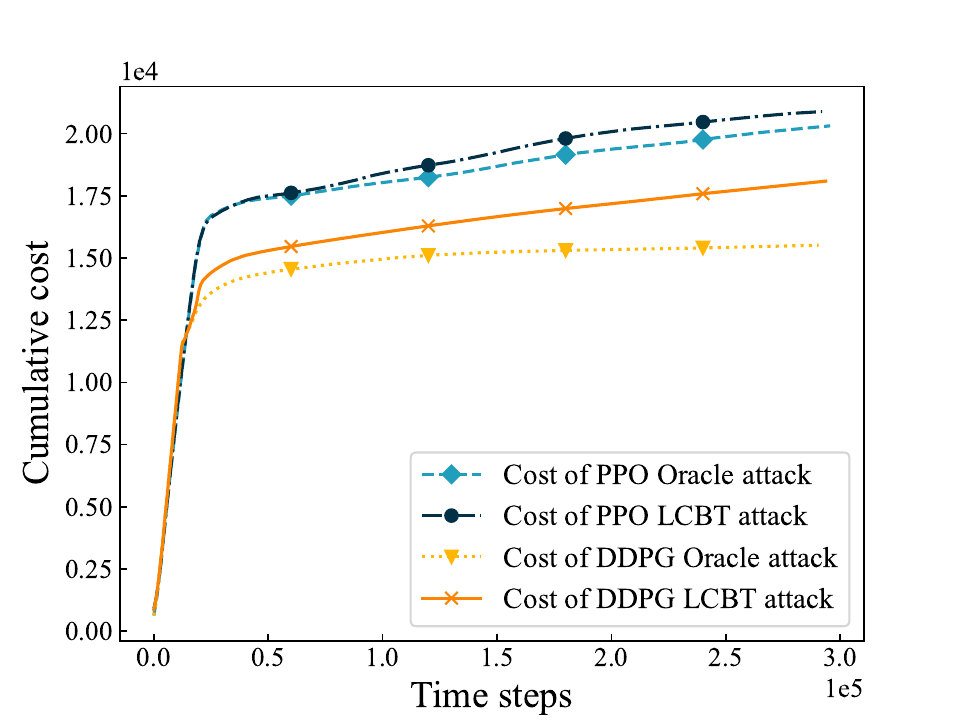}}
\caption{Reward and attack cost results of Environment $1$. In this experiment, $H=10$, $T=3 *10^5$, $r_a = 0.0625$, the corresponding $\K$ for both PPO and DDPG algorithms is 0. In the LCBT attack, the state subspace quantity $M=16$, and $\rho=1/2$. 
}\label{environment1}
\vspace{-6mm}
\end{figure*}

\begin{figure*}[!t]
\centering
\subfigure[Attack DDPG]{
\includegraphics[width=0.32\columnwidth]{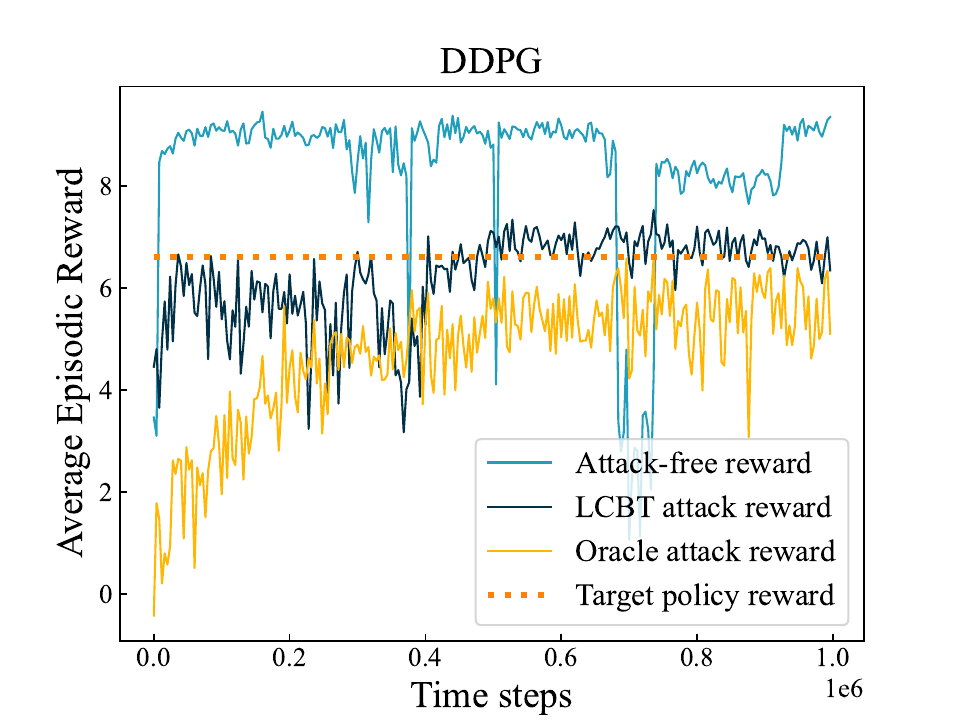}}
\hfill
\subfigure[Attack TD3]{
\includegraphics[width=0.32\columnwidth]{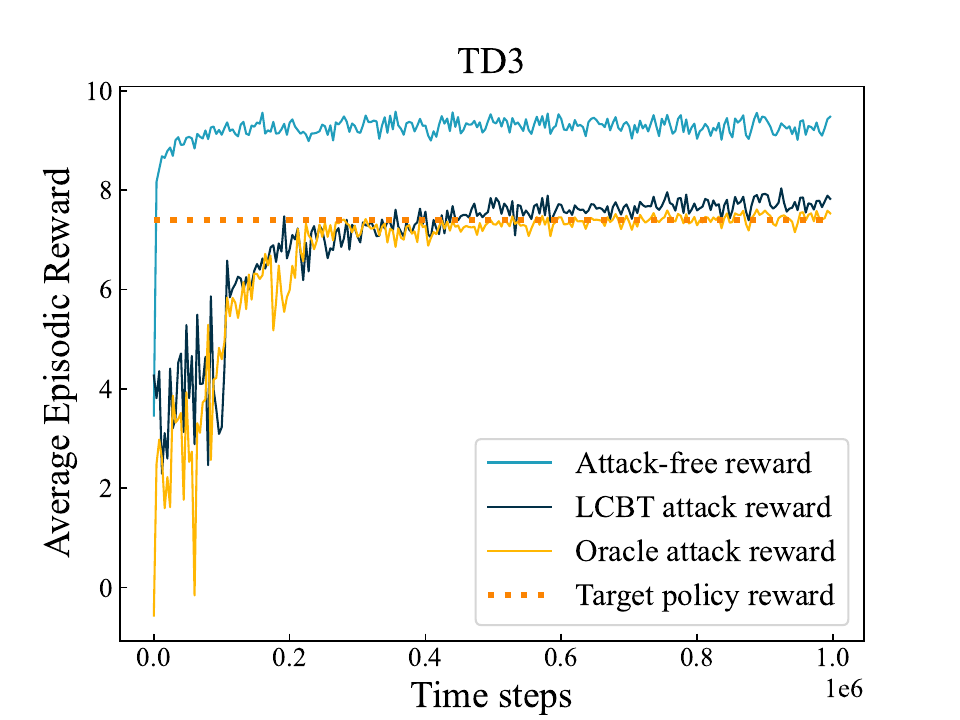}}
\hfill
\subfigure[Attack Cost]{
\includegraphics[width=0.32\columnwidth]{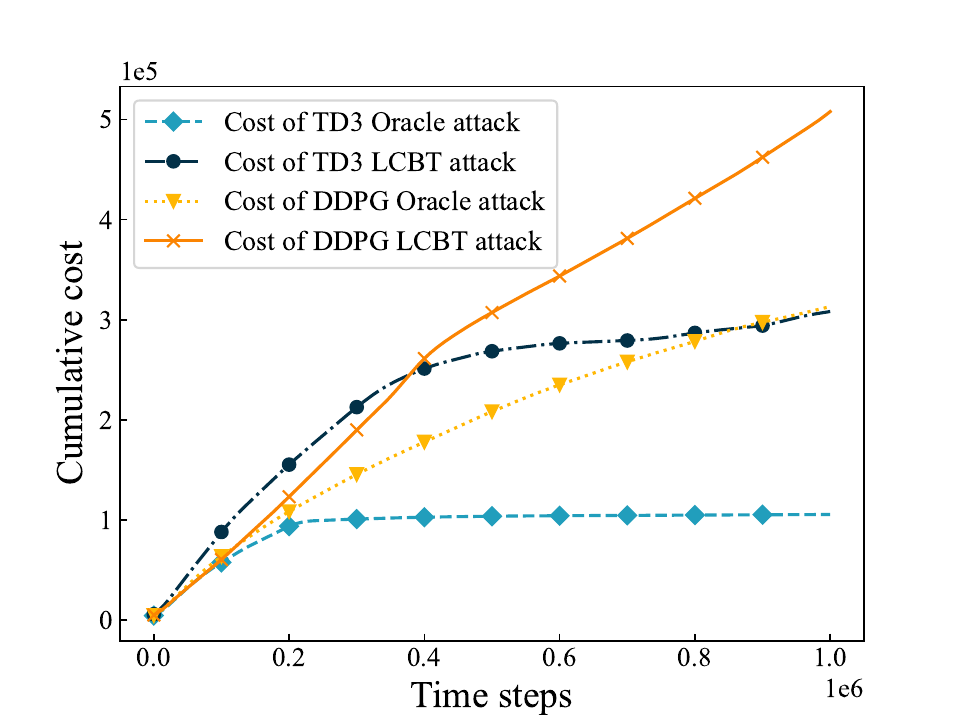}}
\caption{Reward and attack cost results of Environment $2$. In this experiment, $H=10$, $T=10^6$, $r_a = 0.31$, the corresponding $\K$ for both DDPG and TD3 algorithms is 0, $M=81$, and $\rho = 1/\sqrt{2}$.}\label{environment2}
\vspace{-6mm}
\end{figure*}

\textbf{Training Phase:} In Fig. \ref{environment1} (a, b) and Fig. \ref{environment2} (a, b), the target policy reward (\textcolor{orange}{- - -}) is the average reward achieved by the $\pi^{\dagger}$ over $10^3$ episodes. The convergence of rewards obtained by the agent under three different conditions (attack-free, LCBT attack, and oracle attack) is represented by the remaining three curves. Experimental results demonstrate that the reward of the trained policy converges to the reward of the target policy under the oracle attack and the LCBT attack. 
In Fig. \ref{environment1} (c) and Fig. \ref{environment2} (c), 
the cost is sublinear to time steps in both the oracle attack and the LCBT attack, which is in line with our theoretical expectations. Meanwhile, as the attack progresses, after some exploration, the effectiveness of LCBT will also approach that of the oracle attack. The experimental results reveal that the LCBT attack and oracle attack cause the agent to learn policies near the $\pi^{\dagger}$ with a sunlinear attack cost, which indicates the effectiveness of our attack algorithm. Furthermore, from Fig. \ref{environment2}(c), it can be seen that the corresponding attack cost of TD3 is smaller, indicating that TD3 algorithm has a stronger ability to deal with non-stationary environments, resulting in a smaller D-Regret.



\textbf{Test Phase:} In order to further verify the efficiency of our attack, at test time, we measured the percentage of actions in the target action space executed by the agent's trained policy in identical states as a metric for quantifying the similarity between the trained and target policies (The state is generated by the target policy). 
The results are shown in Table \ref{env1} and Table \ref{env2}. From the tables, we can observe that the trained policies shows a high degree of similarity with the target policy. 
Especially under the conditions of the agent using PPO and TD3, the similarity between the trained policies and the target policy can exceed 98\%. In the same environment, different algorithms and corresponding D-Regret can have an impact on the results.


\begin{table}[!htbp]
\vspace{-4mm}
\centering
 \caption{\textup{The similarity between the original target policy and trained policy in environment 1. We have selected policies trained for $3*10^{5}$ steps and $4*10^{5}$ steps under either Oracle attack or LCBT attack. $r_a=0.0625$. 
 }}
 \resizebox{0.8\textwidth}{!}{
\begin{tabular}{c|c|c|c|c|c}
\toprule
\multirow{4}{*}{Environment 1} & \multirow{2}{*}{Training steps} & \multicolumn{2}{c|}{DDPG} & \multicolumn{2}{c}{PPO} \\
\cline{3-6}
& & Oracle attack & LCBT attack & Oracle attack & LCBT attack \\
\cline{2-6}
& 3e5 & 99.816\% & 99.065\% & 98.581\% & 98.642\% \\
\cline{2-6}
& 4e5 & 99.910\% & 99.146\% & 98.640\% & 99.501\% \\
\bottomrule
\end{tabular}
}
\label{env1}
\vspace{-6mm}
\end{table}

\begin{table}[!htbp]
\centering
 \caption{\textup{The similarity between the original target policy and trained policy in environment 2. We have selected policies trained for $7*10^{5}$ steps and $8*10^{5}$ steps under either Oracle attack or LCBT attack. $r_a=0.31$.
 }}
\resizebox{0.8\textwidth}{!}{
\begin{tabular}{c|c|c|c|c|c}
\toprule
\multirow{4}{*}{Environment 2} & \multirow{2}{*}{Training steps} & \multicolumn{2}{c|}{DDPG} & \multicolumn{2}{c}{TD3} \\
\cline{3-6}
& & Oracle attack & LCBT attack & Oracle attack & LCBT attack \\
\cline{2-6}
& 7e5 & 83.600\% & 72.744\% & 99.601\% & 99.102\% \\
\cline{2-6}
& 8e5 & 77.462\% & 70.679\% & 99.864\% & 98.642\% \\
\bottomrule
\end{tabular}
}
\label{env2}
\vspace{-4mm}
\end{table}

\section{Conclusion}

In this study, we investigated the action-manipulation attack to disrupt RL in continuous state and action spaces. We modeled the attack under continuous settings and introduced the concepts of the target action space and the target policy set. Based on the knowledge level of the attacker, we studied white-box and black-box attacks, and provided theoretical guarantees for the efficiency of the attacks. For the design of the black-box attack method, we innovatively use Monte Carlo tree search for action search in order to construct efficient attacks. 



\bibliographystyle{plain}
\medskip

{
\small
\bibliography{ref}
}

\newpage
\appendix


\section{Notation Table}
\label{notationtable}

\begin{table}[!htbp]
\centering
 \caption{Notation Table}
 \resizebox{0.9\textwidth}{!}{
 \begin{tabular}{|c|c|}
 \hline
 \textbf{Notation} & \textbf{Meaning} \\
\hline
 $S$ & The state space \\
 \hline
 $A$ & The action space \\
 \hline
 $H$ & The number of total steps in each episode \\
\hline
 $\mu$ & The initial state distribution \\
 \hline
 $K$ & The number of total episodes \\
 \hline
 $s^k_h$ & The state at step $h$ in episode $k$ \\
 \hline
 $a^k_h$ & The action taken by the agent at step $h$ in episode $k$ \\
 \hline
 $\widetilde{a}^k_h$ & \makecell{The action taken by the attacker and submitted to the environment \\ at step $h$ in episode $k$} \\
 \hline
 $r^k_h$ & The reward generated at step $h$ in episode $k$ \\
 \hline
 $\pi^{\dagger}$ & The target policy specified by the attacker \\
 \hline
 $b^k$ & The behavior policy of episode $k$ \\
 \hline
 $\pi^{\dagger}_h(s)$ & The target action for state $s$ at step $h$\\
 \hline
 $r_a$ & The radius specified by the attacker \\
 \hline
 $\mathcal{A}^{\dagger}_h(s)$ & The target action space, which is defined as $\{a: l(a, \pi_h^{\dagger}(s)) \leq r_a\}$ \\
 \hline
 $\tau$ & The step set whose element is the step when the attacker launches the attack \\
 \hline
 $\alpha$ & \makecell{The step set whose element is the
step when action $a^k_h$ does not belong to \\ the target action space $\mathcal{A}^{\dagger}_h(s^k_h)$} \\
 \hline
 $\Pi^{\dagger}$ & \makecell{The set of target policies that produce actions within the target action space and \\ are superior to policy $\pi^{\dagger}$}\\
 \hline
 $\pi^o$ & The optimal policy in the policy set $\Pi$ \\
 \hline
 $V^{\dagger}_h(s)$ & The value which is equivalent to $V^{\pi^{\dagger}}_h(s)$ \\
 \hline
 $Q^{\dagger}_h(s,a)$ & The value which is equivalent to $Q^{\pi^{\dagger}}_h(s,a)$ \\
 \hline
 $V^o_h(s)$ & The value which is equivalent to $V^{\pi^o}_h(s)$ \\
 \hline
 $Q^o_h(s,a)$ & The value which is equivalent to $Q^{\pi^o}_h(s,a)$ \\
 \hline
 $a^-_h(s)$ & \makecell{The worst action under policy $\pi^o$ and state $s$ at step $h$, \\ which is defined as $\mathop{\arg\min}_{a \in A}Q^o_h(s,a)$} \\
 \hline
 $\Delta_{min}$ & \makecell{The minimum gap between the policy $\pi^o$ and the worst action, \\ which is defined as $\min_{h\in [H], s\in S}\left(V^o_h(s) - Q^o_h(s,a^-_h(s))\right)$} \\
 \hline
 $\K$ & \makecell{The number of episodes in which the attacker does not launch \\ any attacks at the beginning} \\
 \hline
 $l_s$ & The dissimilarity function of the state space\\
 \hline
 $l_a$ & The dissimilarity function of the action space\\
 \hline
 $diam_s(D)$ & The largest difference of all the states in set $D \subseteq S$ w.r.t $l_s$ \\
 \hline
 $diam_a(P)$ & The largest difference of all the actions in set $P \subseteq A$ w.r.t $l_a$ \\
 \hline
 $\T^h_k$ & \makecell{The action cover tree that has been built by the attacker \\ until the beginning of episode $k$ w.r.t step $h$} \\
 \hline
 $P^h_k$ & The traverse path at step $h$ in episode $k$ \\
 \hline
 $|\T^h_k|$ & The node number of tree $\T^h_k$ \\
 \hline
 $\nu_1$ & The maximum distance w.r.t. $l_a$ in the action space \\
 \hline
 $\rho$ & The reduction ratio of nodes in $\T^h_k$ at different depth \\
 \hline
 $(D,I)$ & The node at depth $D$ with index $I$ in action cover trees \\
 \hline
 $\mathcal{P}_{D,I}$ & The subset covered by node $(D,I)$ in the action space $A$ \\
 \hline
 $a_{D,I}$ & The representative action of node $(D,I)$ \\
 \hline
 $M$ & The number of total subintervals of the state space \\
 \hline
 $S_m$ & A subinterval of the state space with $m \in [1,M]$ \\
 \hline
 $i(s)$ & The number of the subinterval to which state $s$ belongs, i.e., $s \in S_{i(s)}$. \\
 \hline
 $d_s$ & The diameter of subintervals of the state space\\
 \hline
 $L_s$ & The coefficient of $d_s$ \\
 \hline
 $\phi^h_{D,I}(k)$ & \makecell{The set of episodes in which the current state belonged to subinterval $S_{i(s^k_h)}$ and node $(D,I)$ \\ was selected by the attacker at the step $h$ until the beginning of the episode $k$} \\
 \hline
 $\gamma^h_{D,I}(k)$ & \makecell{The latest episode before $k$ in which the current state belonged to the subinterval $S_{i(s^k_h)}$ \\ and node $(D,I)$ was selected in the step $h$} \\
 \hline
 $T_{D, I}^h(k)$ & The number of elements in set $\phi^h_{D,I}(k)$. \\
 \hline
 $\hat{Q}^h_{D,I}(s,a_{D,I})$ & The calculated value used to estimate $Q^o_h(s, a_{D,I})$ \\ 
 \hline
 $L^h_{D,I}(k)$ & The lower confidence bound of node $(D,I)$ \\
 \hline
 $B^h_{D,I}(k)$ & The tighter lower confidence bound of node $(D,I)$ \\
 \hline
 
 \end{tabular}
 }
 \label{table:notation}
 \end{table}

\section{Limitations and Broader Impacts}
\label{limitationandimpact}


\textbf{Limitations:} The black-box attack method proposed in this paper depends on the binary tree to divide the action space, which might result in a slight performance loss in a high-dimensional environment. But the setting of parameters, such as $\nu_1$, $\rho$, $L_s$, $d_s$, as well as the RL algorithm used by the agent and the specific environment will all have an impact on the attack cost (e.g., the experiment on Environment 3). 
In addition, if the attacker chooses a low-value target policy, it will lead to an increase in attack cost, which is also reflected in the conclusion.

\textbf{Broader Impacts:} This paper investigates the action-manipulation attack on RL in continuous state and action spaces, introduces the white-box and black-box attack algorithms, and verifies the threat of action-manipulation. We believe that this work has positive implications for the research on robust RL algorithms.

\section{Related works}
\label{relatedwork}

We elucidate the work related to the security of RL from the perspectives of different attack methods.

\textbf{Observation manipulation:} \cite{foley2022execute} employs observation manipulation to cause agent misbehavior only at specific target states. \cite{zhang2021robust}, \cite{yang2020enhanced}, and \cite{pan2019characterizing} minimize the total rewards of the intelligent agent using the observation manipulation method. Moreover, \cite{behzadan2017vulnerability} utilizes the method of observation manipulation to achieve policy induction attacks. \citep{lin2017tactics} aims at minimizing the agent's reward or luring the agent to a designated target state through observation manipulation.

\textbf{Reward manipulation:} \cite{ma2019policy}, \cite{zhang2020adaptive}, \cite{ma2018data} and \cite{huang2019deceptive} employ reward manipulation to train the agent to learn a specified policy, while \cite{majadas2021disturbing} focuses on minimizing the total rewards of the agent using this attack approach. In the study by \cite{wu2022reward}, reward tampering attacks were implemented in multi-agent RL, thus inducing multiple agents to follow the target policy.

\textbf{Environment manipulation:} \cite{xu2021transferable} train agents to use a target policy through environment manipulation. \cite{tanev2021adversarial} and \cite{huang2017adversarial} minimize the total rewards of the agent through environment manipulation. In \cite{boloor2019simple}, the environment is manipulated to cause a controller failure in autonomous driving.


\section{The Oracle Attack Algorithm}
\label{oraclealg}

\begin{algorithm}[htbp]
\caption{The Oracle attack algorithm.}\label{alg:oracle}
\begin{algorithmic}
\REQUIRE ~~\\ 
 		Target policy $\pi^\dag$.
\FOR{episode $k = 1, 2, \dots, K$}
 		\FOR{step $h = 1, 2, \dots, H$}
 		\STATE The agent chooses an action $a_h^k$.
 		\IF {$k \leq \K$ or $a_h^k \in \mathcal{A}^{\dagger}_h(s_h^k)$}
 		\STATE The attacker does not attack, i.e., $\widetilde{a}_h^k =a_h^k$.
 		\ELSE 
            \STATE Set $\widetilde{a}^k_h=a^-_h(s_h^k)$. 
 		\ENDIF
 		\STATE The environment receives action $\widetilde{a}_h^k$, and returns the reward $r_h^k$ and the next state $s_{h+1}^k$.
        \ENDFOR
\ENDFOR
\end{algorithmic}
\label{alg1}
\end{algorithm}

\section{Assumption}
\label{lcbtasm}

\begin{asm}
\label{asm:constant}
We assume that there exist constants $\nu_1$, $0<\rho<1$, $L_s$ and $d_s$ such that:

\begin{itemize}
\item[(a)] 
$diam_a(\mathcal{P}_{D,I}) \leq \nu_1 \rho^D$ for all node $(D,I)$,
\item[(b)] $L_sd_s < \Delta_{min}/2$, and for all $S_m, \forall m \in [1,M]$: $diam_s(S_m) \leq L_sd_s$, 
\item[(c)] $\forall a,a' \in A, s \in S, h \in [1, H]$, $\left|Q^o_h(s,a)-Q^o_h(s,a')\right| \leq l_a(a, a')$,
\item [(d)] $\forall S_i$, $\forall s,s' \in S_i, a \in A, h \in [1, H]$, $\left|Q^o_h(s,a)-Q^o_h(s',a)\right| \leq l_s(s, s')$.
\end{itemize}

\end{asm}

\begin{remark}
    For $(a)$, as the depth increases, the subset of the action space contained in the nodes becomes smaller, and the dissimilarity among actions also decreases, thus, we can always find such constants. For $(b)$, because $\Delta_{min}>0$, there exists a state partitioning method that satisfies the above assumption. $(c)$ and $(d)$ impose a constraint on the smoothness of the $Q$-function concerning state and action, as the black-box attack algorithm involves analysis on the level of action clustering. Consequently, the good smoothness of the $Q$-function corresponding to the target policy helps implement the black-box attack (The setting of the target policy has a direct impact on $\pi^o$).
\end{remark}

\section{Experimental Settings}
\label{expset}

\begin{figure}[htbp]
    \centering
    \includegraphics[width=0.5\linewidth]{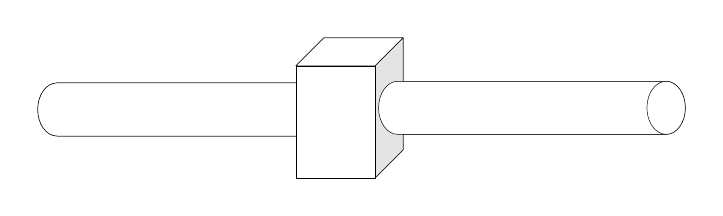}
    \caption{Environment $1$. The objective of this environment is to control the slider to slide on the rod.}
    \label{e1}
\end{figure}

\textbf{Environment $1$.} The objective of this environment, as depicted in Fig. \ref{e1}, is to control the slider to slide on the rod and maximize the reward within $H=10 $ steps. Specifically, for any $h \in [H]$, the state space $s \in \left[-1, 1\right]$ represents the current position of the slider on the rod, where $0$ represents the center point of the rod. The action space $a \in \left[-1, 1\right]$ represents the base sliding distance, with negative values signifying leftward movement and positive values signifying rightward movement. \textcolor{black}{The final sliding distance $d = a*2$}, and a reward of $\left|a\right|$ is obtained for each step taken. If the slider falls off the rod (i.e., the slider's position coordinate is outside of $\left[-1, 1\right]$), the round of the game ends immediately and a reward of $-1$ is received. The slider's initial position $s_1$ is arbitrarily set between $-0.7$ and $0.7$. Our target policy $\pi^{\dagger}$ for the slider is to move only within the interval $[-0.7, 0.7]$. We generate the target policy model by training and obtaining the optimal policy within the limited interval $[-0.7, 0.7]$ through augmentation of the environment constraints. The optimal policy is then employed to act as the target policy. Clearly, $\pi^{\dagger}$ is neither a globally optimal policy nor the worst policy.

\begin{figure}[htbp]
    \centering
    \includegraphics[width=0.4\linewidth]{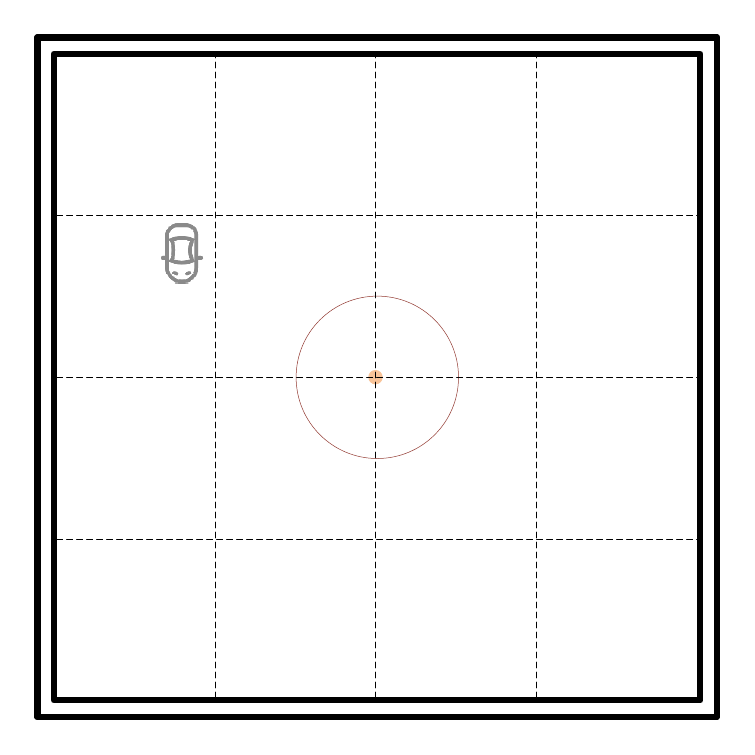}
    \caption{Environment $2$. The objective of this environment is to control a vehicle to move on a two-dimensional plane.}
    \label{e2}
\end{figure}

\textbf{Environment $2$.} As depicted in Fig. \ref{e2}, the objective of this environment is to control a vehicle to move on a two-dimensional plane surrounded by a boundary and obtain maximum reward within $H=10$ steps. For any $h \in [H]$, the state space of Environment $2$ includes the current two-dimensional position of the vehicle, denoted as $\textit{\textbf{s}} = (s_1, s_2) \in \left[0,8\right]^2$. The action space, denoted as $\textit{\textbf{a}} = (a_1, a_2) \in \left[-1,1\right]^2$, denotes the distance of the vehicle's movement. The vehicle's next position is $(s_1+a_1, s_2+a_2)$ and the reward gained is negatively linearly proportional to the distance $d$ between the vehicle and the center point (the yellow dot at $[4, 4]$). The closer the distance, the higher the reward. The initial position $s_1$ of the vehicle is at any point outside the red circle ($d > 1$). Our target policy $\pi^{\dagger}$ for the vehicle is to move only outside the red circle. Similarly, we train and obtain the optimal policy that only moves outside the red circle by increasing the environmental constraints, which serves as the target policy. 

\textbf{Environment $3$.} Environment $3$ is a five-dimensional version of environment $2$, where the goal remains the same: to control an object to move as close as possible to the center point within $H=10$ steps to obtain maximum reward. The state space is denoted as $\textit{\textbf{s}} = (s_1, s_2,s_3,s_4,s_5) \in \left[0,8\right]^5$, the action space as $\textit{\textbf{a}} = (a_1, a_2,a_3,a_4,a_5) \in \left[-1,1\right]^5$, and the next state is given by $(s_1+a_1,s_2+a_2,s_3+a_3,s_4+a_4,s_5+a_5)$. The center point is located at $[4, 4, 4, 4, 4]$. 
The attacker's target policy $\pi^{\dagger}$ is defined as getting as close to the center point as possible without entering the range less than 1 unit distance from the center point.

\section{Additional Numerical Results}
\label{additionalexp}

\begin{figure*}[!t]
\centering
\subfigure[Attack DDPG]{
\includegraphics[width=0.32\columnwidth]{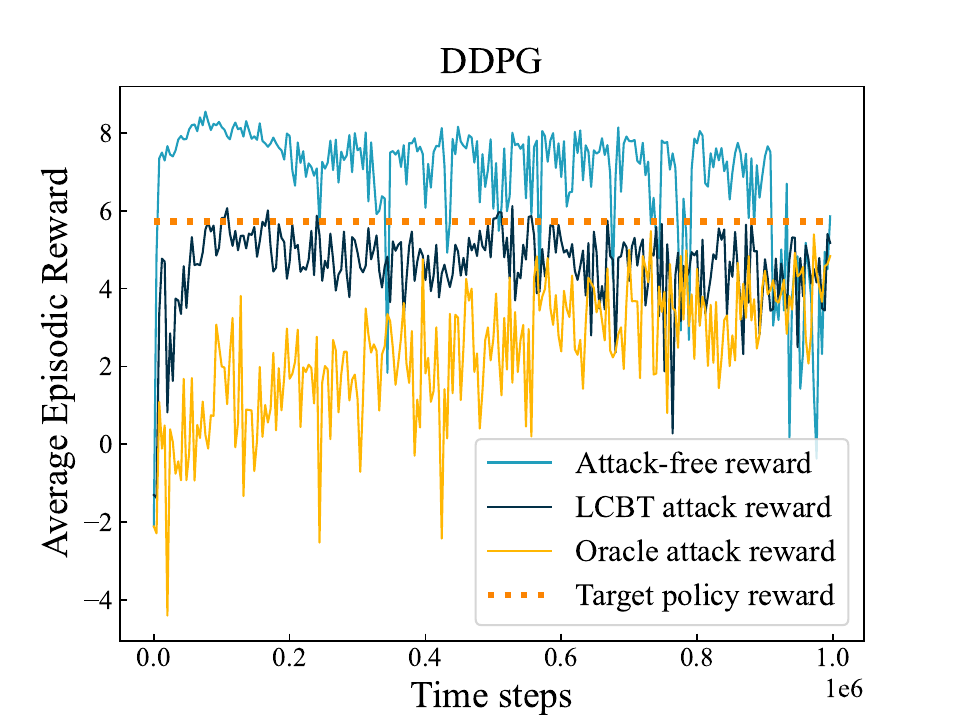}}
\subfigure[Attack TD3]{
\includegraphics[width=0.32\columnwidth]{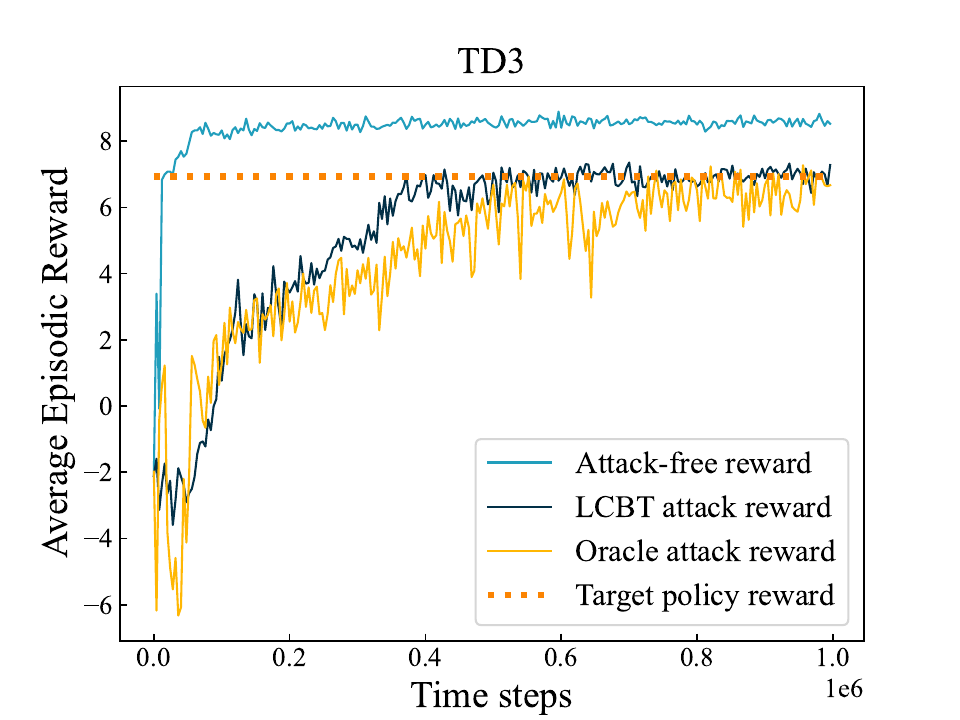}}
\subfigure[Attack Cost]{
\includegraphics[width=0.32\columnwidth]{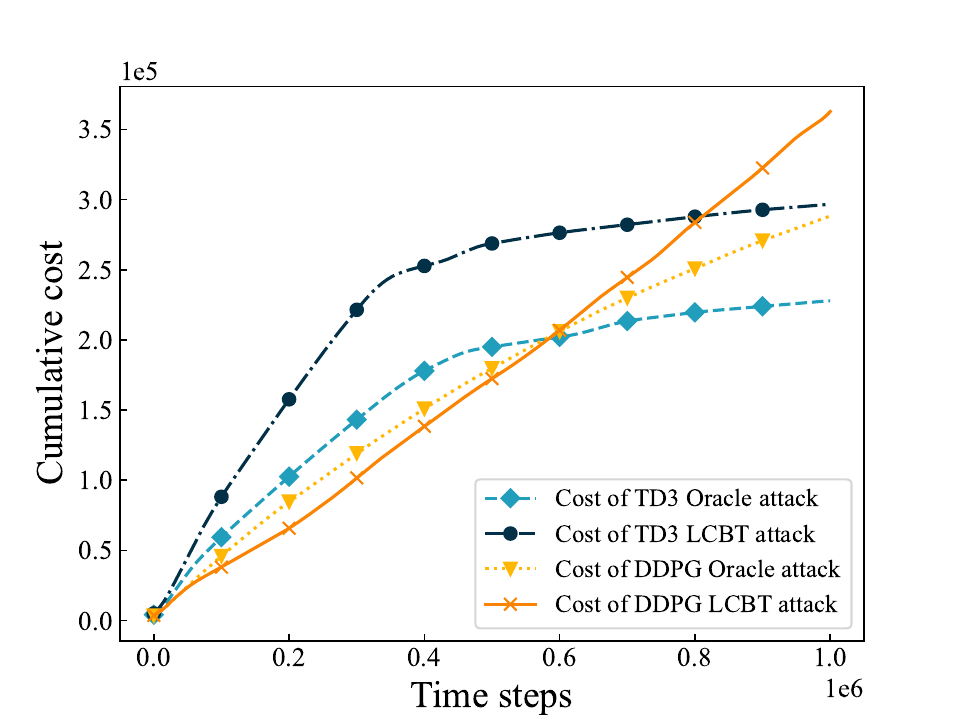}}
\caption{Reward and attack cost results of Environment $3$. In this experiment, $H=10$, $T=10^6$, $r_a = 0.497$, $\K$ for DDPG is 50, $\K$ for TD3 is 100, $M=59049$, and $\rho = 1/\sqrt[5]{2}$.
}\label{environment3}
\end{figure*}


In these three tables, the similarity calculation formula we adopt is as follows: 
\begin{equation*}
sim=\frac{\sum (\mathbb{I}\left \{ l(a_1,a_2) \leq r_a \right \} )}{steps}
\end{equation*}
The $steps$ refers to the total number of steps during the similarity testing period. Here, $steps=1.0*10^{5}$. $a_1$ represents the action output of the attacked policy, while $a_2$ represents the action output of the target policy. $l(\cdot, \cdot)$ denotes the distance between the two actions, where the Euclidean distance is used. For the indicator function $\mathbb{I}\{\xi\}$, if event $\xi$ is established $\mathbb{I}\{\xi\}=1$, otherwise $\mathbb{I}\{\xi\}=0$.


\begin{table}[!htbp]
\centering
 \caption{\textup{The similarity between the original target policy and attacked policy in environment 3. We have selected attacked policies trained for $7*10^{5}$ steps and $8*10^{5}$ steps under either Oracle attack or LCBT attack. $r_a=0.497$. 
 }}
\resizebox{0.8\textwidth}{!}{
\begin{tabular}{c|c|c|c|c|c}
\toprule
\multirow{4}{*}{Environment 3} & \multirow{2}{*}{Training steps} & \multicolumn{2}{c|}{DDPG} & \multicolumn{2}{c}{TD3} \\
\cline{3-6}
& & Oracle attack & LCBT attack & Oracle attack & LCBT attack \\
\cline{2-6}
& 7e5 & 82.502\% & 52.809\% & 95.882\% & 97.039\% \\
\cline{2-6}
& 8e5 & 88.053\% & 64.630\% & 98.251\% & 97.394\% \\
\bottomrule
\end{tabular}
}

\label{env3}
\end{table}

\section{The Proof of Lemma \ref{lem:pre}}
\label{prooflemma1}

In the action-manipulation settings, the attacker sits between the agent and the environment. We can regard the combination of the attacker and the environment as a new environment. For the new environment, we represent the $Q$-value and $V$-value as $\Q$ and $\V$. Because the attacker does not launch an attack when the agent chooses an action within the target action space, we have
\begin{equation*}
    \V_h^o(s) = V_h^o(s) \geq V_h^{\pi}(s) = \V_h^{\pi}(s), \forall s \in S, h \in [1,H], \pi \in \Pi^{\dagger}. 
\end{equation*}

Suppose from the agent's perspective, at the step $h+1$, $\pi^o$ is the optimal policy. At step $h$, the learner takes the action $a \notin \mathcal{A}^{\dagger}_h(s)$, and the attacker manipulates it to $a^-_h(s)$. According to the Bellman equation, we have 
\begin{align*}
\label{eq:QV}
    &\Q_h^*(s,a) = r_h(s, a^-_h(s)) + P_h\V_{h+1}^*(s,a^-_h(s)) \\
    &= r_h(s, a^-_h(s)) + P_h\V_{h+1}^o(s,a^-_h(s)) \\
    &= r_h(s, a^-_h(s)) + P_hV_{h+1}^o(s,a^-_h(s)) \\
    &= Q_h^o(s, a^-_h(s)) \overset{(\romannumeral1)}{<} V_h^o(s) = Q_h^o(s, \pi_h^o(s)) = \Q_h^o(s, \pi_h^o(s))
\end{align*}
where $(\romannumeral1)$ is because $\Delta_{min} > 0$. Therefore, we can conclude that if policy $\pi^o$ is the optimal policy at step $h+1$, then from the perspective of the agent, policy $\pi^o$ remains the optimal policy at step $h$. Since $V_{H+1}^{\pi} = 0$ and $Q_{H+1}^{\pi} = 0$, policy $\pi^o$ is the optimal policy, from induction on $h = H, H-1, ..., 1$. In conclusion, the proof is completed.

\section{Proof of Theorem \ref{theorem:oracle}}
\label{prooftheorem1}

Define $\overline{\Delta}^k_h = \overline{V}^o_h(s^k_h) - \overline{Q}^o_h(s^k_h, a^k_h)$. From Lemma \ref{lem:pre} and the definition of $\pi^o = \sup_{\pi \in \Pi^{\dagger}} V^{\pi}_h(s),\forall s,h$, in the observation of the agent under the oracle attack, $\pi^o$ is the optimal policy. Assuming that $\K = 0$, then the regret of the agent's performance can be defined as
\begin{align}
    \operatorname{Regret}(K) = \sum\limits_{k=1}^K\left[\V^*_1(s^k_1) - \V^{\pi^k}_1(s^k_1)\right] = \sum\limits_{k=1}^K\left[\V^o_1(s^k_1) - \V^{\pi^k}_1(s^k_1)\right], 
\end{align}
where $\pi^k$ is the policy followed by the agent for each episode $k$.

For episode $k$, 
\begin{align*}
&\overline{V}^o_1(s^k_1) - \overline{V}^{\pi^k}_1(s^k_1) \\
&= \overline{V}^o_1(s^k_1) - \mathbb{E}_{a\sim\pi^k_1(\cdot|s^k_1)}\left[\overline{Q}^o_1(s^k_1,a)\right] + \mathbb{E}_{a\sim\pi^k_1(\cdot|s^k_1)}\left[\overline{Q}^o_1(s^k_1,a)\right] - \overline{V}^{\pi^k}_1(s^k_1) \\
&= \mathbb{E}\left[\overline{\Delta}^k_1|\mathcal{F}^k_1\right] + r^k_1 + \mathbb{E}_{s'\sim P_1(\cdot|s^k_1,a\sim\pi^k_1(\cdot|s^k_1))}\overline{V}^o_2(s') - \left(r^k_1 + \mathbb{E}_{s'\sim P_1(\cdot|s^k_1,a\sim\pi^k_1(\cdot|s^k_1))}\overline{V}^{\pi^k}_2(s')\right) \\
&= \mathbb{E}\left[\overline{\Delta}^k_1|\mathcal{F}^k_1\right] + \mathbb{E}_{s'\sim P_1(\cdot|s^k_1,a\sim\pi^k_1(\cdot|s^k_1))}\left[\overline{V}^o_2(s') - \overline{V}^{\pi^k}_2(s')\right] \\
&= ... =\mathbb{E}\left[\sum_{h=1}^H \overline{\Delta}^k_h|\mathcal{F}^k_1\right]
\end{align*}
where $\mathcal{F}^k_h$ represents the $\sigma$-field generated by all the random variables until episode $k$, step $h$ begins. So there exists
\begin{equation}
    \sum\limits_{k=1}^K \left(\V^o_1(s^k_1) - \V^{\pi^k}_1(s^k_1)\right) 
    = \mathbb{E}\left[\sum\limits^K_{k=1}\sum\limits^H_{h=1}\overDelta^k_h|\mathcal{F}^k_1\right].
\end{equation}

Next, we will show that with a probability at least $1-\delta_2$, we have
\begin{align}
\label{martigale}
    \sum\limits^K_{k=1}\sum\limits^H_{h=1}\overDelta^k_h \leq \sum\limits_{k=1}^K \left(\V^o_1(s^k_1) - \V^{\pi^k}_1(s^k_1)\right) + 2H^2 \sqrt{\ln(1/ \delta_2)\sum_{k=1}^K \left(\V^o_1(s^k_1) - \V^{\pi^k}_1(s^k_1)\right)}. 
\end{align}
Since $\mathbb{E}\left[\sum_{h=1}^H \overDelta^k_h|\mathcal{F}^k_1\right] = \V^o_1(s^k_1) - \V^{\pi^k}_1(s^k_1)$, we can regard 
\begin{equation}
Y_k:=\sum_{i=1}^k\left(\sum_{h=1}^H \overDelta^i_h - \left(\V^o_1(s^i_1) - \V^{\pi^k}_1(s^i_1)\right)\right),  
\end{equation}
as a martingale with the difference sequence $\{X_k\}_{k=1}^K$, which is
\begin{equation}
    X_k := \sum_{h=1}^H\overDelta^k_h - \left(\V^o_1(s^k_1) - \V^{\pi^k}_1(s^k_1)\right). 
\end{equation}

And we have the difference sequence bounded, i.e., $|X_k| \leq H^2, \forall k \in [1, K]$. Define the predictable quadratic variation process of the martingale: $W_K := \sum_{k=1}^K \mathbb{E}[X_k^2|\mathcal{F}^k_1]$, with
\begin{align}
     W_K &\leq \sum_{k=1}^K \mathbb{E}\left[\big(\sum_{h=1}^H\overDelta^k_h\big)^2|\mathcal{F}^k_1\right] \leq H^2 \sum_{k=1}^K \mathbb{E}\left[\sum_{h=1}^H\overDelta^k_h|\mathcal{F}^k_1\right] \nonumber \\ 
     &= H^2 \sum_{k=1}^K \left( \V^o_1(s^k_1) - \V^{\pi^k}_1(s^k_1) \right) = \sigma^2. 
\end{align}

By Freeman's inequality \citep{10.1214/ECP.v16-1624}, we have
\begin{align*}
    &\mathbb{P}\left(Y_K=\sum_{k=1}^K X_k > 2H^2 \sqrt{\ln(1/\delta_2)\sum_{k=1}^K\big(\V^o_1(s^k_1) - \V^{\pi^k}_1(s^k_1)\big)} \right) \\
    \leq & \exp\left\{ -\frac{4H^4\ln(\frac{1}{\delta_2})\sum\limits_{k=1}^K\big(\V^o_1(s^k_1) - \V^{\pi^k}_1(s^k_1)\big)/2}{\sigma^2 + H^2 \cdot 2H^2 \sqrt{\ln(\frac{1}{\delta_2})\sum\limits_{k=1}^K\big(\V^o_1(s^k_1) - \V^{\pi^k}_1(s^k_1)\big)} / 3} \right\} \\
    \leq& \exp\{-\ln(1/\delta_2)\} = \delta_2,
\end{align*}
and combine the definition of $Y_k$, we can get (\ref{martigale}). 

Under the oracle attack, when the agent chooses an action satisfying $a^k_h \in \mathcal{A}^{\dagger}_h(s^k_h)$, the attacker does nothing, and we have $\Q^o_h(s^k_h,a^k_h) = Q^o_h(s^k_h,a^k_h) \leq V^o_h(s^k_h)$. Otherwise, the attacker launches the attack, and the action $a^k_h$ will be replaced by $a^-_h(s^k_h)$. That is to say, we have $\Q^o_h(s^k_h,a^k_h) = r_h(s^k_h,a^-_h(s^k_h)) + P_h\V^o_{h+1}(s^k_h,a^-_h(s^k_h)) = r_h(s^k_h,a^-_h(s^k_h)) + P_hV^o_{h+1}(s^k_h,a^-_h(s^k_h)) = Q^o_h(s^k_h, a^-_h(s^k_h))$. Then, we can obtain 
\begin{align*}
    &\sum_{k=\K + 1}^K \sum_{h=1}^H \overDelta^k_h \\
    &= \sum_{k=\K + 1}^K \sum_{h=1}^H \left(\V^o_h(s^k_h) - \Q^o_h(s^k_h,a^k_h)\right) \\
    &= \sum_{k=\K + 1}^K \sum_{h=1}^H \left(V^o_h(s^k_h) - \Q^o_h(s^k_h,a^k_h)\right) \\
    &= \sum_{k>\K, (k,h) \notin \alpha} V^o_h(s^k_h) - Q^o_h(s^k_h,a^k_h) + \sum_{k>\K, (k,h) \in \alpha} V^o_h(s^k_h) - Q^o_h(s^k_h, a^-_h(s^k_h)) \\
    &\geq \sum_{k>\K, (k,h) \in \alpha} V^o_h(s^k_h) - Q^o_h(s^k_h, a^-_h(s^k_h)) \\
    &\geq \sum_{k>\K, (k,h) \in \alpha} \Delta_{min} \\
    &\geq |\zeta| \Delta_{min}. 
\end{align*}
where $\zeta$ is a set defined as $\{(k,h):k > \K, (k,h) \in \alpha\}$.

With $|\zeta| = |\tau|$, we have 
\begin{align*}
    |\tau| \leq \frac{\sum_{k=\K+1}^K \sum_{h=1}^H \overDelta^k_h}{\Delta_{min}} \overset{(\romannumeral1)}{\leq} \frac{\Regret(K) + 2H^2\sqrt{\ln(1/\delta_2)\cdot \Regret(K)}}{\Delta_{min}},  
\end{align*}
and with $|\zeta| \geq |\alpha| - H\K$, we have $|\alpha| \leq |\tau| + H\K$, 
$(\romannumeral1)$ is obtained by (\ref{martigale}).

\section{Proof of Lemma \ref{confidencebound}}
\label{prooflemma2}

\begin{lemma}
\label{confidencebound}
    In the LCBT algorithm, the following confidence bound:
    \begin{equation}
        \Big|\hat{Q}^h_{D,I}(s^k_h,a_{D,I}) - \mathbb{E}[\hat{Q}^h_{D,I}(s^k_h,a_{D,I})]\Big| \leq \frac{H-h+1}{\sqrt{2T^h_{D,I}(k)}}\sqrt{\ln \left(\frac{2Mk\sum_{h=1}^H|\T^h_k|}{\delta_1}\right)}
    \end{equation}
    holds for $\forall h\in [H], S_m (s^k_h \in S_m, m \in [M]), (D,I) \in \T^k_h, T^h_{D,I}(k) \in [1,k]$ with a probability at least $1-\delta_1$.
\end{lemma}
The lemma offers a confidence bound for $\hat{Q}^h_{D,I}(s^k_h,a_{D,I})$. According to the Algorithm \ref{wortraverse}, the attacker traverses the tree with smaller $B$-values, which means $\hat{Q}^h_{D,I}(s^k_h,a_{D,I})$ will converge to $Q^o_h(s^k_h,a^-_h(s^k_h))$.

\textbf{Proof:} Firstly, we will transform $\hat{Q}$-value into a non-recursive form, i.e.,
\begin{align*}
    &\hat{Q}^h_{D,I}(s^k_h,a_{D,I}) \\
    &= (1-\frac{1}{T^h_{D,I}(k)})\hat{Q}^h_{D,I}(s^{\gamma^h_{D,I}(k)}_h,a_{D,I}) + \frac{1}{T^h_{D,I}(k)} (r_h^k + G^k_{h+1:H+1} \cdot \rho^k_{h+1:H+1}) \\
    &= \frac{T^h_{D,I}(k)-1}{T^h_{D,I}(k)}\Big[(1-\frac{1}{T^h_{D,I}(k)-1})\hat{Q}^h_{D,I}(s^{\gamma^h_{D,I}(\gamma^h_{D,I}(k))}_h,a_{D,I}) \\
    &\qquad + \frac{1}{T^h_{D,I}(k)-1}(r_h^{\gamma^h_{D,I}(k)} + G^{\gamma^h_{D,I}(k)}_{h+1:H+1} \cdot \rho^{\gamma^h_{D,I}(k)}_{h+1:H+1})\Big] \\ 
    &\qquad + \frac{1}{T^h_{D,I}(k)}(r_h^k + G^k_{h+1:H+1} \cdot \rho^k_{h+1:H+1})\\
    &= ... = \frac{1}{T^h_{D,I}(k)}\sum\limits_{i=1}^k\mathbb{I}\{s^i_h \in S_{i(s^k_h)}, a^i_h = a_{D,I}\}(r^i_h + G^i_{h+1:H+1} \cdot \rho^i_{h+1:H+1}). 
\end{align*}
Then we have 
\begin{align}
    &\mathbb{E}\left[\hat{Q}^h_{D,I}(s^k_h,a_{D,I})\right] \nonumber \\
    &= \frac{\sum\limits^k_{i=1} \mathbb{I}\{s^i_h \in S_{i(s^k_h)}, a^i_h = a_{D,I}\} \mathbb{E}\left[r^i_h + G^i_{h+1:H+1} \cdot \rho^i_{h+1:H+1}\right]}{T^h_{D,I}(k)} \nonumber \\
    &= \frac{1}{T^h_{D,I}(k)} \sum\limits^k_{i=1} \mathbb{I}\{s^i_h \in S_{i(s^k_h)}, a^i_h = a_{D,I}\} Q^o_h(s^i_h, a^i_h) \label{Q:average} \\
    &= Q^o_h(s^0_{k,h}, a_{D,I}) \nonumber
\end{align}
It should be noted that for $\mathbb{E}[\hat{Q}^h_{D,I}(s^k_h,a_{D,I})]$, there exists a state $s^0_{k,h} \in S_{i(s^k_h)}$ such that $Q^o_h(s^0_{k,h},a_{D,I})=\mathbb{E}[\hat{Q}^h_{D,I}(s^k_h,a_{D,I})]$ holds due to the smoothness of $Q$-function (Assumption \ref{asm:constant} (d)).

Define the event:
\begin{align*}
    \xi_k &= \bigg\{\forall h\in [H], m \in [M], (D,I) \in \T^h_k, T^h_{D,I}(k) \in [1,k], \\ 
    &\left| \hat{Q}^h_{D,I}(s^k_h,a_{D,I})-\mathbb{E}\left[\hat{Q}^h_{D,I}(s^k_h,a_{D,I})\right] \right| \leq \beta^h_k\left(T^h_{D,I}(k),\delta_1\right) \bigg\},
\end{align*}
where $\beta$-function is calculated by 
\begin{equation}
    \beta^h_k\left(N,\delta\right) = \frac{H-h+1}{\sqrt{2N}}\sqrt{\ln\left(\frac{2Mk\sum^H_{h=1}|\T^h_k|}{\delta}\right)}. 
\end{equation}

Define $\xi_k^c$ as the opposite event of $\xi_k$, then we have 
\begin{align*}
    &\mathbb{P}(\xi^c_k) = \sum\limits_{h=1}^H \sum\limits_{m=1}^M \sum\limits_{(D,I) \in \T^h_k} \sum\limits_{T^h_{D,I}(k)=1}^k \\
    &\quad \mathbb{P}\left(\left| \hat{Q}^h_{D,I}(s^k_h,a_{D,I})-\mathbb{E}\left[\hat{Q}^h_{D,I}(s^k_h,a_{D,I})\right] \right| > \beta^h_k(T^h_{D,I}(k), \delta_1)\right) \\ 
    &\overset{(\romannumeral1)}{<} \sum\limits_{h=1}^H \sum\limits_{m=1}^M \sum\limits_{(D,I) \in \T^h_k} \sum\limits_{T^h_{D,I}(k)=1}^k 2 \exp\left[\frac{2\beta^h_k(T^h_{D,I}(k),\delta_1)^2 T^h_{D,I}(k)^2}{-T^h_{D,I}(k)(H-h+1)^2} \right] \\
    &= \sum\limits_{h=1}^H \sum\limits_{m=1}^M \sum\limits_{(D,I) \in \T^h_k} \sum\limits_{T^h_{D,I}(k)=1}^k 2 \exp\left[-\frac{2\beta^h_k(T^h_{D,I}(k),\delta_1)^2 T^h_{D,I}(k)}{(H-h+1)^2} \right] \\
    &= \sum\limits_{h=1}^H \sum\limits_{m=1}^M \sum\limits_{(D,I) \in \T^h_k} \sum\limits_{T^h_{D,I}(k)=1}^k 2\exp \left[-\ln\left(\frac{2Mk\sum_{h=1}^H|\T^h_k|}{\delta_1}\right)\right] \\
    &= \sum\limits_{h=1}^H \sum\limits_{m=1}^M \sum\limits_{(D,I) \in \T^h_k} \sum\limits_{T^h_{D,I}(k)=1}^k 2 \cdot \frac{\delta_1}{2Mk\sum_{h=1}^H|\T^h_k|} = \delta_1,  
\end{align*}
where $(\romannumeral1)$ is utilizing the Hoeffding's inequality. So there exists $\mathbb{P}(\xi_k) \geq 1-\delta_1$, the proof is completed. 

\section{Proof of Lemma \ref{meantomin}}
\label{prooflemma3}

\begin{lemma}
\label{meantomin}
    In the LCBT algorithm, based on the lemma \ref{confidencebound}, there exists
    \begin{equation}
        \mathbb{E}[\hat{Q}^h_{D,I}(s^k_h,a_{D,I})] - Q^o_h(s^k_h, a^-_h(s^k_h)) \leq 3 \cdot \frac{H-h+1}{\sqrt{2T^h_{D,I}(k)}}\sqrt{\ln \left(\frac{2Mk\sum_{h=1}^H|\T^h_k|}{\delta_1}\right)} + L_sd_s. 
    \end{equation}
\end{lemma}
The Lemma gives the relationship between $\mathbb{E}[\hat{Q}^h_{D,I}(s^k_h,a_{D,I})]$ and $Q^o_h(s^k_h, a^-_h(s^k_h))$. 

\textbf{Proof:} From the traverse function, along the path $P^h_k$, we have 
\begin{align}
    B^h_{D,I}(k) &= \max\left[L^h_{D,I}(k), \min\limits_{j \in \{2I-1, 2I\}} B^h_{D+1,j}(k)\right] \label{calB}\\
    & \geq \min\limits_{j\in \{2I-1,2I\}} B^h_{D+1,j}(k) \nonumber\\
    &= B^h_{D+1,I'}(k)\left((D+1, I') \in P^h_k\right). \nonumber 
\end{align}

We make an assumption that at the step $h$ in episode $k$, the attacker launches the attack and chooses a node $(D,I)$ along the path $P^h_k$, then we have 
\begin{equation}
    B^h_{D',I'}(k) \geq B^h_{D,I}(k) \geq L^h_{D,I}(k) \big(D'<D, (D',I') \in P^h_k\big). 
\end{equation}

Because the root node includes $a^-_h(s^k_h)$, so in the path $P^h_k$, except $(D,I)$, there must exist a node $(D_{min},I_{min})(D_{min} < D)$ containing action $a^-_h(s^k_h)$. So we have 
\begin{equation}
\label{BminL}
    B^h_{D_{min},I_{min}}(k) \geq B^h_{D,I}(k) \geq L^h_{D,I}(k), 
\end{equation}
established. We now show that for any node $(D_m,I_m)$ such that $a^-_h(s^k_h) \in \mathcal{P}_{D_m,I_m}$, then $Q^o_h(s^k_h, a^-_h(s^k_h))$ is a valid upper bound of $L^h_{D_m,I_m}(k)$. From the definition of $L^h_{D,I}(k)$, we can obtain 
\begin{align*}
    &L^h_{D_{m},I_{m}}(k) \\
    &= \hat{Q}^h_{D_{m},I_{m}}(s^k_h, a_{D_{m},I_{m}}) - \beta^h_k\left(T^h_{D_{m},I_{m}}(k), \delta_1\right) - L_sd_s - \nu_1\rho^{D_{m}} \\
    &\overset{(\romannumeral1)}{\leq} \mathbb{E}\left[\hat{Q}^h_{D_{m},I_{m}}(s^k_h, a_{D_{m},I_{m}})\right] - L_sd_s - \nu_1\rho^{D_{m}} \\
    &\overset{(\romannumeral2)}{=} Q^o_h(s^0_{k,h},a_{D_{m},I_{m}}) - L_sd_s - \nu_1\rho^{D_{m}} \\
    &\overset{(\romannumeral3)}{\leq} Q^o_h(s^k_h, a_{D_{m},I_{m}}) - \nu_1\rho^{D_{m}} \\
    &\overset{(\romannumeral4)}{\leq} Q^o_h(s^k_h, a^-_h(s^k_h))
\end{align*}
where $(\romannumeral1)$ is under Lemma \ref{confidencebound}, $(\romannumeral2)$ is according to (\ref{Q:average}), $(\romannumeral3)$ is under Assumption \ref{asm:constant} (b, d), and $(\romannumeral4)$ is under Assumption \ref{asm:constant} (a, c). There exists a leaf node $(D^-,I^-)$ containing action $a^-_h(s^k_h)$, then 
\begin{equation*}
    B^h_{D^-,I^-}(k) = L^h_{D^-,I^-}(k) \leq Q^o_h(s^k_h, a^-_h(s^k_h)),
\end{equation*}
and obviously, all the nodes containing action $a^-_h(s^k_h)$ with $D > D_{min}$ are descendants of $(D_{min},I_{min})$. Now by propagating the upper bound of nodes containing $a^-_h(s^k_h)$, i.e., $Q^o_h(s^k_h, a^-_h(s^k_h))$, backward from $(D^-,I^-)$ to $(D_{min}, I_{min})$ through (\ref{calB}), we can show that $Q^o_h(s^k_h, a^-_h(s^k_h))$ is a valid upper bound of $B^h_{D_{min},I_{min}}$. 

Then from (\ref{BminL}), we have, i.e.,
\begin{align}
    &Q^o_h(s^k_h, a^-_h(s^k_h)) \geq B^h_{D_{min},I_{min}}(k) \geq B^h_{D,I}(k) \geq L^h_{D,I}(k) \nonumber\\
    &= \hat{Q}^h_{D,I}(s^k_h,a_{D,I}) - \beta^h_k(T^h_{D,I}(k), \delta_1) - L_sd_s - \nu_1\rho^D \nonumber \\
    &\overset{(\romannumeral1)}{\geq} \mathbb{E}\left[\hat{Q}^h_{D,I}(s^k_h,a_{D,I})\right] - 2\beta^h_k(T^h_{D,I}(k), \delta_1) - L_sd_s - \nu_1\rho^D \nonumber \\
    &\overset{(\romannumeral2)}{\geq} \mathbb{E}\left[\hat{Q}^h_{D,I}(s^k_h,a_{D,I})\right] - 3\beta^h_k(T^h_{D,I}(k), \delta_1) - L_sd_s, \label{mintodi}
\end{align}
$(\romannumeral1)$ is under Lemma \ref{confidencebound}, and $(\romannumeral2)$ is because the selected node is always a leaf node, which satisfies $\beta^h_k(T^h_{D,I}(k), \delta_1) > \nu_1\rho^D$. 

Based on (\ref{mintodi}), we can obtain
\begin{align}
    \mathbb{E}\left[\hat{Q}^h_{D,I}(s^k_h,a_{D,I})\right] - Q^o_h(s^k_h, a^-_h(s^k_h)) \leq 3\beta^h_k(T^h_{D,I}(k), \delta_1) + L_sd_s. 
\end{align}
The proof is completed.

\section{Proof of Theorem \ref{theorem:lcbt}}
\label{prooftheorem2}

Since the attacker will not launch the attack when the agent chooses an action satisfying $a^k_h \in \mathcal{A}^{\dagger}_h(s^k_h)$, we have $\Q^o_h(s^k_h,a^k_h) = Q^o_h(s^k_h,a^k_h) \leq V^o_h(s^k_h)$. Otherwise, the attacker selects a node $(D,I)$ according to the LCBT algorithm, and replaces $a^k_h$ to the corresponding action $a_{D,I}$, i.e., $\Q^o_h(s^k_h,a^k_h) = r_h(s^k_h,a_{D,I}) + P_h\V^o_{h+1}(s^k_h,a_{D,I}) = r_h(s^k_h,a_{D,I}) + P_hV^o_{h+1}(s^k_h,a_{D,I}) = Q^o_h(s^k_h,a_{D,I})$. Then we have
\begin{align}
    &\sum_{k=\K+1}^K \sum_{h=1}^H \overDelta^k_h = \sum_{k=\K+1}^K \sum_{h=1}^H \left(\V^o_h(s^k_h) - \Q^o_h(s^k_h,a^k_h)\right) \nonumber \\
    &= \sum_{k=\K+1}^K \sum_{h=1}^H \left(V^o_h(s^k_h) - \Q^o_h(s^k_h,a^k_h)\right) \nonumber\\
    &= \sum_{k>\K, (k,h) \notin \alpha} V^o_h(s^k_h) - Q^o_h(s^k_h,a^k_h) \nonumber + \sum_{k>\K, (k,h) \in \alpha} V^o_h(s^k_h) - Q^o_h(s^k_h,a_{D,I}) \nonumber \\
    &\overset{(\romannumeral1)}{\geq} \sum_{(k,h) \in \zeta} V^o_h(s^k_h) - \left(Q^o_h(s^0_{k,h}, a_{D,I}) + L_sd_s\right) \nonumber \\
    &\overset{(\romannumeral2)}{=} \sum_{(k,h) \in \zeta} V^o_h(s^k_h) - \mathbb{E}\left[\hat{Q}^h_{D,I}(s^k_h,a_{D,I})\right] - L_sd_s \nonumber \\
    &\overset{(\romannumeral3)}{\geq} \sum_{(k,h) \in \zeta} V^o_h(s^k_h) - Q^o_h(s^k_h, a^-_h(s^k_h)) - 3\beta^h_k(T^h_{D,I}(k), \delta_1) - 2 L_sd_s \nonumber \\
    &\geq |\zeta|(\Delta_{min} - 2 L_sd_s) - \sum_{(k,h) \in \zeta} 3\beta^h_k(T^h_{D,I}(k), \delta_1),  
    \label{deltabeta}
\end{align}
where $\zeta=\{(k,h):k > \K, (k,h) \in \alpha\}$, $(\romannumeral1)$ is under Assumption \ref{asm:constant} (b, d), $(\romannumeral2)$ is according to (\ref{Q:average}) and $(\romannumeral3)$ is under Lemma \ref{meantomin}. Next, we analyze $\sum_{(k,h) \in \alpha} \beta^h_k(T^h_{D,I}(k), \delta_1)$. Define $N_{D,I}^h(S_m):=|\{\gamma:\gamma \leq K, s^\gamma_h \in S_m, a^\gamma_h=a_{D,I}\}|$ as the total number of episodes that in step $h$, the corresponding state belongs to $S_m$ and the attacker selects node $(D,I)$, and further
\begin{align}
&\sum\limits_{(k,h) \in \zeta } {\frac{1}{{\sqrt {T_{D,I}^h(k)} }}}  = \sum\limits_{h = 1}^H {\sum\limits_{m = 1}^M {\sum\limits_{(D,I) \in \T_K^h} {\sum\limits_{N = 1}^{N_{D,I}^h({S_m})} {\frac{1}{{\sqrt N }}} } } } \nonumber \\
 &\leq \sum\limits_{h = 1}^H {\sum\limits_{m = 1}^M {\sum\limits_{(D,I) \in \T_K^h} {\int_1^{N_{D,I}^h({S_m})} {\frac{1}{{\sqrt N }}{dN}} } } } \nonumber \\
 &\le \sum\limits_{h = 1}^H {\sum\limits_{m = 1}^M {\sum\limits_{(D,I) \in \T_K^h} {2\sqrt {N_{D,I}^h({S_m})} } } } \nonumber \\
 &= M \cdot \sum\limits_{h = 1}^H {\left| {\T_K^h} \right|}  \cdot \frac{1}{{M \cdot \sum\limits_{h = 1}^H {\left| {\T_K^h} \right|} }}\sum\limits_{h = 1}^H {\sum\limits_{m = 1}^M {\sum\limits_{(D,I) \in \T_K^h} {2\sqrt {N_{D,I}^h({S_m})} } } } \nonumber \\
 &\overset{(\romannumeral1)}{\le} 2M \cdot \sum\limits_{h = 1}^H {\left| {\T^h_K} \right|} \sqrt {\frac{{\sum_{h = 1}^H {\sum_{m = 1}^M {\sum_{(D,I) \in \T_K^h} {N_{D,I}^h({S_m})} } } }}{{M \cdot \sum_{h = 1}^H {\left| {\T_K^h} \right|} }}} \nonumber \\
 &= 2 \sqrt {M \sum_{h = 1}^H {\left| {\T_K^h} \right|}  \cdot \left| \zeta  \right|}.
\label{boundT}
\end{align}
where $(\romannumeral1)$ is based on the property of the concave function $\sqrt{n}$.

Then combine (\ref{deltabeta}), we have
\begin{align}
&\sum_{k=\K+1}^K \sum_{h=1}^H \overDelta^k_h 
\ge |\zeta|(\Delta_{min} - 2 L_sd_s) - 3 \sum_{(k,h) \in \zeta} \frac{H-h+1}{\sqrt{2T^h_{D,I}(k)}}\sqrt{\ln\left(\frac{2Mk\sum^H_{h=1}|\T^h_k|}{\delta_1}\right)} \nonumber\\
&\overset{(\romannumeral1)}{\ge} |\zeta|(\Delta_{min} - 2 L_sd_s) - 3H \cdot \sqrt {\ln \left( {\frac{{2MK \cdot HK}}{{{\delta_1}}}} \right)}  \cdot \sqrt {2 M \sum\limits_{h = 1}^H {\left| {\T_K^h} \right| }\cdot\left| \zeta  \right| } \label{deltalower}
\end{align}
where $(\romannumeral1)$ is obtained based on (\ref{boundT}). Use $(\ref{martigale})$ to bound $\sum^K_{k=1}\sum^H_{h=1}\overDelta^k_h$, i.e., 
\begin{align}
   &\sum\limits^K_{k=\K+1}\sum\limits^H_{h=1}\overDelta^k_h \overset{(\romannumeral1)}{\leq} \sum\limits^K_{k=1}\sum\limits^H_{h=1}\overDelta^k_h \nonumber \\
   &\leq \sum\limits_{k=1}^K \left(\V^o_1(s^k_1) - \V^{\pi^k}_1(s^k_1)\right) + 2H^2 \sqrt{\ln(1/ \delta_2)\sum_{k=1}^K \left(\V^o_1(s^k_1) - \V^{\pi^k}_1(s^k_1)\right)} \nonumber\\
    &\overset{(\romannumeral2)}{\leq} \sum\limits_{k=1}^K \left(\V^{\pi^{*,k},k}_1(s^k_1) - \V^{\pi^k,k}_1(s^k_1)\right) + 2H^2 \sqrt{\ln(1/ \delta_2)\sum_{k=1}^K \left(\V^{\pi^{*,k}, k}_1(s^k_1) - \V^{\pi^k,k}_1(s^k_1)\right)} \nonumber\\
    & = \operatorname{D-Regret}(K) + 2H^2 \sqrt{\ln(1/\delta_2) \operatorname{D-Regret}(K)}, \label{deltaupper} 
\end{align}
where $\pi^{*,k} = \sup_{\pi}V^{\pi,k}_1(s^k_1)$ is the optimal policy of episode $k$. In $(\romannumeral1)$, we assume $\K=0$. The reason for $(\romannumeral2)$ is that because of the existence of the attacker, the environment is non-stationary in the observation of the agent. Combine (\ref{deltalower}) and (\ref{deltaupper}), we can obtain
\begin{align}
    &\operatorname{D-Regret}(K) + 2H^2 \sqrt{\ln(1/\delta_2) \operatorname{D-Regret}(K)} \ge \nonumber \\ 
    & \left| \zeta  \right|(\Delta_{min}  - 2 {L_s}{d_s} ) - 3\sqrt {\left| \zeta  \right|} \cdot H \cdot \sqrt {\ln \left( {\frac{{2MK \cdot HK}}{{{\delta _1}}}} \right)}  \cdot \sqrt {2M \cdot \sum\limits_{h = 1}^H {\left| {\T_K^h} \right|} }. \nonumber 
\end{align}

Finally, we can obtain
\begin{align}
    \left| \zeta  \right| \le \frac{2\left(\operatorname{D-Regret}(K) + 2H^2 \sqrt{\ln(1/\delta_2) \cdot \operatorname{D-Regret}(K)}\right)}{\Delta_{min} - 2 L_sd_s} + \frac{18 M H^2 \ln(\frac{2MHK^2}{\delta_1}) \sum_{h=1}^H |\T^h_K|}{(\Delta_{min} - 2 L_sd_s)^2},  
\end{align}
with $|\zeta| = |\tau|$ and $|\alpha| \leq |\zeta| + H\K$, the proof is completed.

\section{The Node Number and Height of the Cover Tree} \label{nodenumber}

In this section, we bound the node number of the cover tree, i.e., $|\T^h_K|$ for $\forall h \in [1, H]$. 

A node $(D, I)$ will be expanded when it satisfies the condition
\begin{equation*}
    {\nu_1}{\rho ^D} \ge \frac{{H - h + 1}}{{\sqrt {2T^h_{D,I}(k)} }} \cdot \sqrt {\ln \left( {\frac{{2Mk \cdot \sum_{h = 1}^H {\left| {\T_k^h} \right|} }}{{{\delta _1}}}} \right)}.  
\end{equation*}

We make some transformations, i.e., 
\begin{align*}
    T^h_{D,I}(k) &\ge \frac{{{{(H - h + 1)}^2}}}{{2\nu _1^2{\rho ^{2D}}}} \cdot \ln \left( {\frac{{2Mk \cdot \sum_{h = 1}^H {\left| {\T_k^h} \right|} }}{{{\delta _1}}}} \right) \\
    &\overset{(\romannumeral1)}{>} \frac{{{{(H - h + 1)}^2}}}{{2\nu _1^2{\rho ^{2D}}}} \cdot \ln \left( {\frac{{2M \cdot 3H}}{{{\delta _1}}}} \right) \\
    &= \frac{{{{(H - h + 1)}^2}}}{{2\nu _1^2{\rho ^{2D}}}} \cdot \ln \left( {\frac{{6MH}}{{{\delta _1}}}} \right), 
\end{align*}
$(\romannumeral1)$ is because the minimum number of nodes in $\T^h_K$ is 3. 

From the inequality above, we can see that as the depth $D$ increases, the $T^h_{D,I}(k)$-value which can make the inequality above established will also increase ($0< \rho < 1$), so it's obvious that when the tree $\T^h_k$ is a complete binary tree, its total number of nodes is the most. Based on this, we assume that the depth is $D_m$, i.e., the nodes at depth $1,2,..., D_m-1$ have been expanded, then we have
\begin{align*}
    K &\ge \sum\limits_{D = 1}^{D{}_m - 1} {{2^D} \cdot \frac{{{{(H - h + 1)}^2}}}{{2\nu _1^2{\rho ^{2D}}}} \cdot \ln \left( {\frac{{6MH}}{{{\delta _1}}}} \right)} \\
    &= \sum\limits_{D = 1}^{D{}_m - 1} {{{\left( {2{\rho ^{ - 2}}} \right)}^D} \cdot \frac{{{{(H - h + 1)}^2}}}{{2\nu _1^2}} \cdot \ln \left( {\frac{{6MH}}{{{\delta _1}}}} \right)} \\
    &= \frac{{{{(H - h + 1)}^2}}}{{\nu _1^2}} \cdot \ln \left( {\frac{{6MH}}{{{\delta _1}}}} \right) \cdot \frac{{{{\left( {2{\rho ^{ - 2}}} \right)}^{{D_m} - 1}} - 1}}{{2 - {\rho ^2}}}. 
\end{align*}

Then we can get the upper bound of $D_m$, i.e., 
\begin{equation}
{D_m} \le {\log _{2{\rho ^{ - 2}}}}\left[ {\frac{{K \cdot \nu _1^2 \cdot \left( {2 - {\rho ^2}} \right)}}{{{{(H - h + 1)}^2} \cdot \ln \left( 6MH/\delta_1 \right)}} + 1} \right] + 1. 
\end{equation}

Through $2^{D_m+1} - 1$, we get the upper bound of the node number of tree $\T^h_K$, i.e., 
\begin{align}
\left| {\T_K^h} \right| &\le {4}{\left[ {\frac{{K \cdot \nu _1^2 \cdot \left( {2 - {\rho ^2}} \right)}}{{{{(H - h + 1)}^2} \cdot \ln \left( 6MH/\delta_1 \right)}} + 1} \right]^{{{\log }_{2{\rho ^{ - 2}}}}2}} \nonumber \\
&= O\left( {{K^{^{{{\log }_{2{\rho ^{ - 2}}}}2}}}} \right). 
\end{align}

Note that because $0<\rho<1$, we always have $\log_{2\rho^{-2}} 2 <1$. 
Regarding the height of $\T^h_K$, it is evident that the height is at its maximum when only one node per level of $\T^h_K$ is expanded. Consequently, the upper bound of the height can similarly be derived, which is $O(\log_{\rho^{-2}}K)$.

\begin{figure}[!t]
\centering
\subfigure{
\includegraphics[width=0.32\columnwidth]{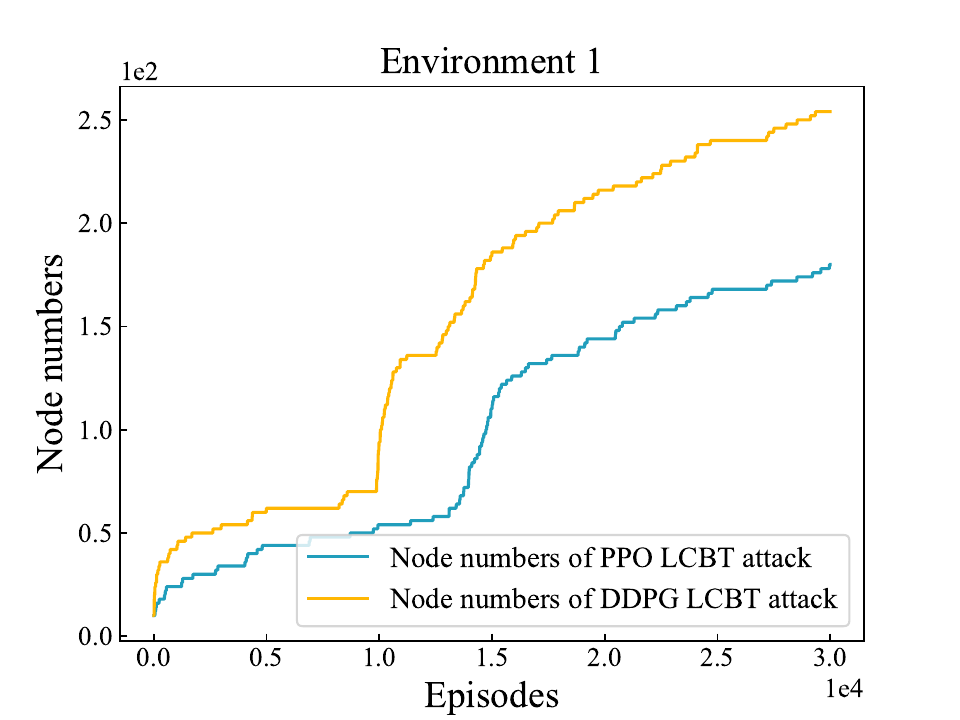}}
\hspace{0in}
\subfigure{
\includegraphics[width=0.32\columnwidth]{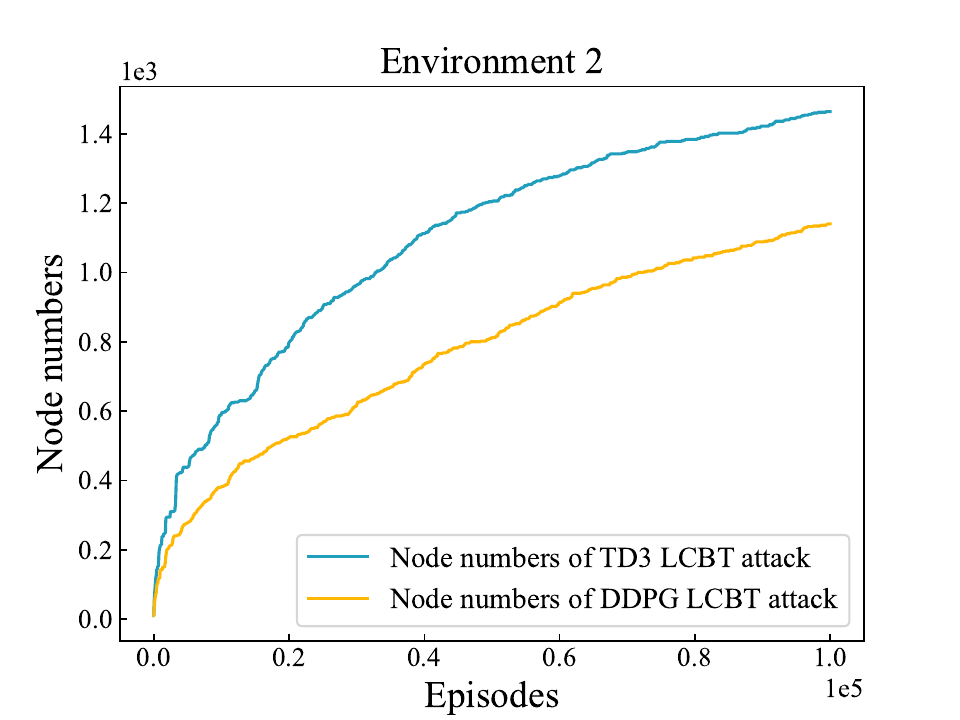}}
\subfigure{
\includegraphics[width=0.32\columnwidth]{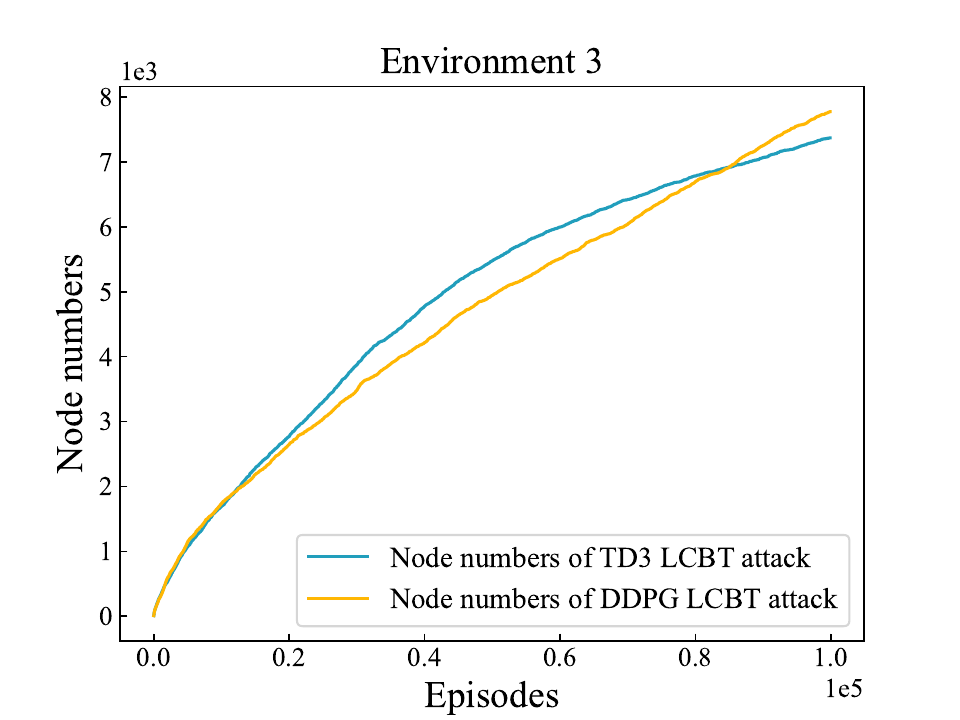}}
\caption{The total number of nodes of the cover trees.}\label{nodes}
\end{figure}


In Fig. \ref{nodes}, the graph shows the relationship between the number of episodes and the total number of nodes of all the cover trees, i.e., $\sum_{h=1}^H |\T^h_k|$. The x-axis indicates the number of episodes during the training phase, while the y-axis represents the total number of nodes of all the cover trees after each corresponding episode. It is evident that compared to the increase in the number of episodes, the increase in the number of nodes is slow, and the final number of nodes is far smaller than the number of total episodes. 

\section{Time and Space Complexity Analysis}
\label{timespace}

\begin{lemma}
    In the LCBT algorithm, the computational time complexity is $O(H\cdot K^{1+E}+H\cdot K \cdot \log_{\rho^{-2}}K)$. 
\end{lemma}

\textbf{Proof:} The computational time complexity of the LCBT algorithm primarily emanates from two aspects: (1) After each round, the attacker needs to update the corresponding $L$ and $B$ values for each node of every tree, and the complexity of this step is related to the number of nodes in the tree. Specifically, assuming that in the $k$-th episode, after completion, the attacker updates each node of $H$ trees, according to the proof in Appendix \ref{nodenumber}, the upper bound on the number of nodes in a tree is $O(k^E)$, where $E=\log_{2\rho^{-2}}2<1$. Therefore, for the total $K$ episodes, the time complexity of this part is $O(K\cdot H\cdot K^E)$. (2) When the attacker launches an attack, it is necessary to traverse from the root node to the leaf node (Algorithm \ref{wortraverse}), and the complexity of this step is related to the depth of the tree. By employing a proof method akin to that in Appendix \ref{nodenumber}, we can ascertain an upper bound for the depth of the tree as $O(\log_{\rho^{-2}}k)$ in episode $k$. Making an extreme assumption that the attacker launches an attack at every step, the time complexity for this part over the total $K$ episodes is $O(K\cdot H \cdot \log_{\rho^{-2}}K)$. In summary, the overall computational time complexity is given by: $O(H\cdot K^{1+E}+H\cdot K \cdot \log_{\rho^{-2}}K)$. 

\begin{figure}[!t]
\centering
\includegraphics[width=0.7\columnwidth]{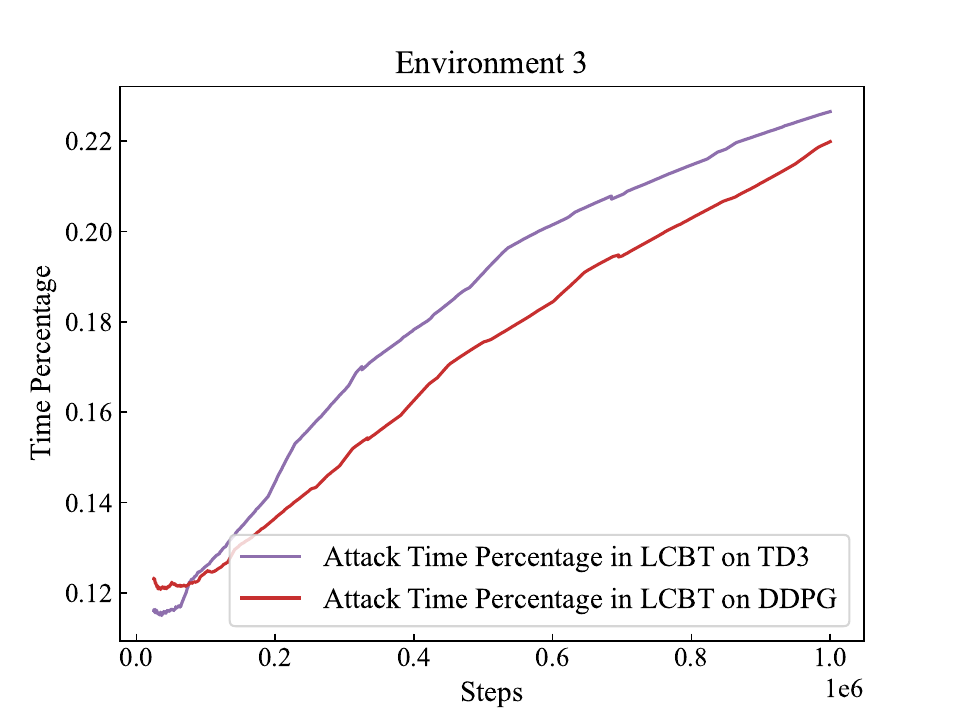}
\caption{The percentage of time taken to execute the LCBT algorithm during the attack phase.}\label{timerecord}
\end{figure}

We recorded the time taken to execute the attack algorithm and the total time during the training phase, calculated the ratio between the two, and obtained the results as depicted in Fig. \ref{timerecord}.

\begin{lemma}
    In the LCBT algorithm, the space complexity is $O(MHK^E)$. 
\end{lemma}

\textbf{Proof:} When the attacker executes the LCBT algorithm, it needs to store $H$ action cover trees, that is, $\sum_{h=1}^H |\T^h_K|$, and each node in the tree corresponds to $M$ states and an action, then it is necessary to store $M$ $\hat{Q}$-values, so the space complexity is $M \cdot \sum_{h=1}^H |\T^h_K| = O(MHK^E)$. 

\end{document}